\documentclass[lettersize,journal]{IEEEtran}
\usepackage{amsthm}
\newtheorem{corollary}{Corollary}

\newtheorem{assumption}[corollary]{Assumption}
\usepackage{amsmath,amsfonts}
\theoremstyle{plain}
\usepackage{algorithmic}
\usepackage{algorithm}
\usepackage{array}
\usepackage[caption=false,font=normalsize,labelfont=sf,textfont=sf]{subfig}
\usepackage{textcomp}
\usepackage{stfloats}
\usepackage{url}
\usepackage{verbatim}
\usepackage{graphicx}
\usepackage{cite}
\usepackage{hyperref}
\usepackage{xcolor}
\usepackage{colortbl, booktabs}
\usepackage{multirow}
\usepackage{booktabs}  
\usepackage{amssymb}   
\usepackage{pifont}    
\usepackage{booktabs}  
\usepackage{amssymb}   
\usepackage{pifont}    
\usepackage{algorithm}
\usepackage{algorithmic}
\newcommand{\xmark}{\ding{55}}  
\begin{document}

\title{Causal Disentanglement-Inspired Degradation Representation Learning for Full-Reference Image Quality Assessment}

\author{
Zhen Zhang,
Jielei Chu,~\IEEEmembership{Senior~Member,~IEEE,}
Tian Zhang,~\IEEEmembership{Member,~IEEE,}
Lin Ma,
Fengmao Lv,~\IEEEmembership{Member,~IEEE,}
Weide Liu,~\IEEEmembership{Member,~IEEE,}
Tianrui Li,~\IEEEmembership{Senior~Member,~IEEE,}
Yuming Fang,~\IEEEmembership{Fellow,~IEEE}
\thanks{
Zhen Zhang, Jielei Chu, Fengmao Lv, and Tianrui Li are with the School of Computing
and Artificial Intelligence, Southwest Jiaotong University, Chengdu 611756,
China (e-mail: \{zhenzhang, jieleichu, fengmaolv, trli\}@swjtu.edu.cn).

Lin Ma is with the School of Transportation and Logistics, Southwest Jiaotong University,
Chengdu, China (e-mail: malin@swjtu.edu.cn).

Tian Zhang is with the School of Physics, Northeast Normal University,
Changchun 130024, China (e-mail: zhangt100@nenu.edu.cn).

Weide Liu is with the School of Computing and Artificial Intelligence, Jiangxi University of Finance and Economics, Nanchang, 330013, China (e-mail: weide001@e.ntu.edu.sg).

Yuming Fang is with the School of Information Management,
Jiangxi University of Finance and Economics, Nanchang, China
(e-mail: fa0001ng@e.ntu.edu.sg).

\vspace{1em}
Corresponding author: Jielei Chu.
}
}


\maketitle
 
    \begin{abstract} Existing deep network-based Full-Reference Image Quality Assessment (FR-IQA) models typically rely on pairwise comparisons of deep features from reference and distorted images. Although effective on standard IQA benchmarks, existing FR-IQA methods still face two major limitations. Training-dependent methods require labeled IQA data for supervised optimization, but reliable quality annotations are difficult to obtain because they require labor-intensive subjective experiments. Training-free methods avoid supervised training, but their fixed perceptual priors limit their adaptability to non-standard or domain-specific image scenarios. To address these limitations, we propose a novel FR-IQA paradigm inspired by causal disentanglement representation learning. Unlike conventional feature comparison-based FR-IQA methods, our approach reformulates degradation representation learning from a causal disentanglement-inspired perspective. Specifically, degradation and content representations are first decoupled by exploiting the content invariance between reference and distorted images. Then, inspired by the human visual masking effect, a masking module is designed to model the influence of image content on degradation features, thereby extracting content-influenced degradation representations from distorted images. Finally, quality scores are predicted from these representations using either supervised regression or dimensionality reduction. Extensive experiments show that our method achieves highly competitive performance on standard IQA benchmarks under fully supervised, few-shot, and label-free settings. Moreover, our method exhibits stronger label-free adaptation than existing training-free FR-IQA models on diverse non-standard image domains with scarce data, including infrared, radiographic, screen-content, medical, and synthetic images. Code: https://github.com/yuchen152/Causal-Disentanglement-for-Full-Reference-Image-Quality-Assessment

\end{abstract}

\begin{figure}[t] 
	\centering
	\includegraphics[width=0.49\textwidth]{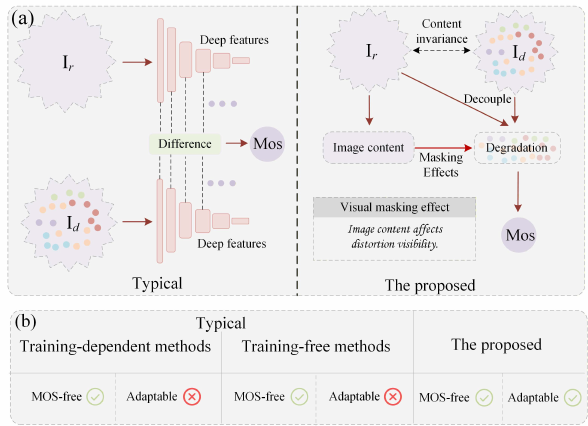}
	\caption{(a) The typical FR-IQA architecture. Most deep FR-IQA models compute the differences between feature maps to predict quality scores. The proposed FR-IQA architecture extracts content-influenced degradation features to predict quality scores. (b) The comparison of FR-IQA paradigms. Training-free methods avoid MOS labels but are difficult to adapt efficiently, while training-dependent methods achieve adaptation through MOS-supervised training. In contrast, our method enables efficient label-free adaptation by learning content-influenced degradation representations.}
	\label{2}
\end{figure}


\section{Introduction}
\IEEEPARstart{I}{mage} Quality Assessment (IQA)~\cite{continualIQA, Structural} is essential for perceptual image processing, image restoration, compression, and multimedia delivery. Therefore, developing IQA methods that align well with human subjective perception has become important. IQA can be categorized into Full-Reference IQA (FR-IQA)~\cite{wang2004image,zhang2011fsim,saha2016full}, Reduced-Reference IQA (RR-IQA)~\cite{rehman2012reduced}, and No-Reference IQA (NR-IQA)~\cite{mittal2012no, GMAD}. FR-IQA models are generally believed to be less practical than NR-IQA models because of their reliance on pristine images. However, recent studies~\cite{ding2020image, DeepJSD} have shown that FR-IQA models can guide perceptual image enhancement, a task that poses challenges for many NR-IQA models~\cite{zhang2022perceptual}. As such, FR-IQA models remain crucial.

FR-IQA estimates perceptual quality by comparing a distorted image with its pristine reference. In previous studies, this difference was usually regarded as a measure of the distance between these images. As shown in Figure~\ref{2}, deep network-based FR-IQA models often employ deep learning for feature decomposition and score pooling. These FR-IQA models typically extract feature maps from both distorted and reference images, and compute their differences to derive quality scores. Deep network-based FR-IQA methods can be categorized into end-to-end deep learning models (training-dependent) and training-free models. Although these methods have achieved great success in the field of FR-IQA, several limitations remain. \textbf{Training-dependent} methods~\cite{ding2020image, chen2024topiq, liao2025image} rely on large-scale labeled IQA data. However, due to the high subjectivity of human perception, obtaining reliable ground truth typically necessitates a large panel of human observers to conduct exhaustive scoring or ranking (e.g., Mean Opinion Score (MOS)). Consequently, this heavy reliance restricts their scalability and applicability, particularly in data-scarce or emerging domains (e.g., infrared or screen-content images). Moreover, \textbf{training-free} methods~\cite{DSD,shen2025image,DeepJSD,DeepDC} typically leverage pretrained models (e.g., VGG~\cite{simonyan2014very}, EfficientNet~\cite{tan2019efficientnet}) trained on ImageNet to extract feature maps from both reference and distorted images. Then, they compute the distance between these maps to derive the quality scores. However, these methods cannot be readily adapted to specific datasets, leading to severe performance drops in some scenarios. For instance, DBIQA~\cite{liao2025image} demonstrated that DeepWSD~\cite{liao2022deepwsd} only achieves excellent performance in limited scenarios. Therefore, it is essential to develop an FR-IQA method that does not rely on subjective quality annotations and can be efficiently adapted to specific datasets. In this work, we propose a method inspired by causal disentanglement representation learning that can be efficiently adapted to diverse scenarios (e.g., image enhancement or radiographic imaging) without the need for costly MOS annotations.

To achieve label-free adaptability, it is necessary to establish unsupervised paradigms for both training and prediction. During training, our method is based on the premise that a distorted image can be viewed as consisting of two components, namely image content and image degradation. In FR-IQA, distorted images are usually generated by applying degradations to reference images. This generation process supports this premise and ensures that the image content is shared between the reference image and the distorted image. By exploiting this content invariance, our method decouples degradation features from image content in a self-supervised manner without MOS annotations. However, decoupling degradation alone is not sufficient to model subjective human perception. According to the visual masking effect (VME), the Human Visual System (HVS) has different sensitivities to the same distortion under different image contents. For example, noise is highly visible in smooth regions, such as a clear sky, but can be nearly imperceptible in textured regions, such as dense grass. This phenomenon indicates that the perception of image degradation depends on the combined influence of image content and distortion. Since image content affects the visibility of degradation, it is necessary to explicitly model how image content modulates the degradation representation. 

To this end, we propose a degradation representation method to extract content-influenced degradation features inspired by causal disentanglement representation learning. During prediction, supervised or few-shot regression can be used to learn an explicit mapping from these features to subjective scores, which further improves the alignment with human visual perception. When label annotations are unavailable, absolute quality scores are not identifiable without additional supervision. Therefore, we formulate the label-free setting as relative quality ranking rather than absolute quality score prediction. This formulation is motivated by the observation that quality-aware representations often contain a dominant low-dimensional structure that is strongly aligned with subjective quality, and that neighboring samples in the feature space tend to exhibit similar perceptual quality. Based on this observation, we use dimensionality reduction methods to map the learned content-influenced degradation feature to a one-dimensional space. The resulting coordinate serves as a relative quality coordinate, enabling label-free quality ranking without MOS supervision.

The main contributions of this work are summarized as follows.
\begin{itemize}
\item Unlike conventional deep-learning-based FR-IQA methods that estimate degradation by measuring feature difference between reference and distorted images, we propose a causal disentanglement perspective for degradation representation in FR-IQA. Specifically, the content invariance of FR-IQA is first exploited to decouple image content from image degradation. Then, we formulate the influence of image content on degradation visibility based on the visual masking effect, which further aligns the learned representation with human subjective perception.

\item We design a complete algorithmic framework to implement the proposed paradigm. The framework integrates degradation encoding, visual masking module (VMM), and degradation-conditioned reconstruction into a unified pipeline. It learns content-influenced degradation representations without MOS annotations.

\item We conduct extensive experiments to validate the effectiveness and generalization ability of the proposed method. On standard IQA benchmarks, the method achieves highly competitive performance under fully supervised, few-shot, and label-free settings. On diverse non-standard image domains, including infrared, radiographic, screen-content, medical, and synthetic images, the method shows stronger label-free performance than existing training-free FR-IQA models, demonstrating its practical value in data-scarce scenarios.

\end{itemize}

The remainder of this paper is structured as follows. Section~\ref{sec:relaywork} reviews the relevant studies. Section~\ref{Problem Formulation} introduces the preliminaries of this work, where we formulate FR-IQA from a causal-disentanglement perspective and describe the underlying representation-learning principle. Section~\ref{Method} presents the proposed framework in detail, including degradation feature decoupling, content-conditioned visual masking, quality score prediction, model architecture, and its connections to the human visual system and existing FR-IQA paradigms. Section~\ref{Experiments} reports the experimental evaluation, covering implementation protocols, comparisons on standard IQA benchmarks, experiments on domain-specific datasets, controlled validation of degradation disentanglement and visual masking, and further analyses. Section~\ref{Conclusions} concludes the paper, and outlines future research directions. Additional theoretical analyses, supplementary ablation studies, and visualization results are provided in the Appendix.

\section{Related Work}
\label{sec:relaywork}

\subsection{Various Quality Assessment Scenarios}

The development of FR-IQA models has been closely tied to practical quality assessment scenarios. Early benchmark datasets, such as TID2013~\cite{ponomarenko2015image}, LIVE~\cite{sheikh2006statistical}, CSIQ~\cite{larson2010most}, and KADID~\cite{lin2019kadid}, were mainly developed to study human perception of fidelity loss caused by synthetic distortions under controlled settings. More recent datasets, including PIPAL~\cite{jinjin2020pipal}, further extend FR-IQA to images generated by modern restoration and synthesis algorithms, where perceptual quality is often harder to determine because algorithm-induced changes do not always correspond to quality degradation. Meanwhile, FR-IQA has also been explored in more specific scenarios~\cite{underwater, EEG, Color}, such as screen-content~\cite{ni2017scid}, infrared~\cite{zelmati2022study}, medical~\cite{medical}, and radiographic images~\cite{zhang2025comprehensive}. However, progress in these areas remains limited, largely because IQA annotation is costly, subjective, and often requires domain expertise, making labeled datasets difficult to obtain. Moreover, existing training-free FR-IQA methods show performance degradation in cross-domain scenarios, since they usually rely on ImageNet-pretrained feature representations that are primarily optimized for natural images. When applied to domain-specific data, the discrepancy in image statistics and structural information leads to domain shift and compromises prediction reliability. These limitations call for an FR-IQA framework that can be adapted to domain-specific features without relying on subjective quality labels.

\subsection{Deep Network-Based FR-IQA Models}

Deep networks have been widely used to model the complex relationship between reference and distorted images in FR-IQA. Early studies extracted sensitivity or similarity maps from both images for quality estimation~\cite{gao2017deepsim,kim2017deep}. Later, researchers found that VGG-based deep features effectively represent perceptual similarity, serving as useful losses for image synthesis and quality prediction~\cite{zhang2018unreasonable}. Building on this, Ding et al.~\cite{ding2020image} proposed DISTS, which measures structural and textural similarity in deep feature space, and ADISTS~\cite{ADISTS} further improves it using a dispersion index for adaptive weighting. Liao et al.~\cite{liao2022deepwsd} instead modeled perceptual degradation using Wasserstein distance in feature space, emphasizing distributional differences for quality evaluation. TOPIQ~\cite{chen2024topiq} introduces a top-down approach featuring a cross-scale attention mechanism, which leverages high-level semantics to guide the network's focus toward important local distortion regions for efficient quality evaluation. DBIQA~\cite{liao2025image} introduces a dual-branch framework to capture joint degradation effects in deep feature space by combining self-similarity analysis with pairwise reference-distorted feature comparison. DeepCausal~\cite{shen2025image} investigates causal perceptual effects in FR-IQA through abductive counterfactual inference, aiming to reveal causal relationships between deep network features and perceptual distortions. DeepDC~\cite{DeepDC} quantifies statistical similarity in the deep feature space using distance correlation, thereby reducing the dependence on strict pixel-level alignment. DeepJSD~\cite{DeepJSD} further explores deep distributional measures, including Jensen-Shannon divergence and symmetric Kullback-Leibler divergence, to compare pretrained network features for training-free FR-IQA.

However, existing methods still face practical constraints. Training-dependent models are limited by the scarcity of annotated IQA data, while training-free models struggle to adapt quickly to domain-specific datasets. Despite these bottlenecks, these methods have achieved considerable success by formulating FR-IQA as a measure of deep feature differences. In this paper, we propose a causal disentanglement-inspired FR-IQA framework that learns degradation representations under the guidance of the visual masking effect.

\subsection{Causal Disentanglement Representation Learning}

Disentangled representation learning aims to separate the underlying generative factors within data into independent and interpretable components. Early approaches, such as Independent Component Analysis (ICA)~\cite{ICA}, attempt to decompose signals into statistically independent sources. Variational Autoencoder (VAE)~\cite{kingma2013auto} further advances this idea by introducing regularization and prior constraints to structure the latent space, while $\beta$-VAE~\cite{higgins2017beta} enhances disentanglement through KL divergence optimization. However, traditional methods often assume independent latent factors, which is unrealistic for data with complex causal dependencies. To address this, CausalVAE~\cite{yang2021causalvae} incorporates causal structures to achieve causal disentanglement, and SCADI~\cite{nam2023scadi} employs pseudo-label generation for semi-supervised causal disentanglement. In this paper, we formulate FR-IQA from the perspective of causal disentanglement representation learning.

\subsection{Visual Masking Effects}

Recent research reveals that image content strongly affects the visibility of distortions during visual perception~\cite{ahumada1992luminance, pelli2013measuring, legge1980contrast, chen2010perceptually}. The perceived visual quality of an image is affected by several factors, including contrast sensitivity function (CSF), luminance adaptation (LA), contrast masking (CM), and foveated masking (FM). Spatial CSF indicates the bandpass properties of the HVS in the spatial frequency domain~\cite{ahumada1992luminance}. Compared with high-frequency regions, the HVS is more sensitive to distortion in low-frequency regions. LA reflects HVS sensitivity as influenced by background luminance~\cite{pelli2013measuring}. It is harder to perceive distortion in relatively dark or bright environments. CM refers to the reduced visibility of one contrast pattern in the presence of another~\cite{legge1980contrast}. FM implies that visual sensitivity is related to retinal eccentricity, with distortion far from the focus of gaze being less perceptible~\cite{chen2010perceptually}. In IQA, visual masking is commonly modeled to mimic human subjective perception~\cite{wang2023visual, zheng2024cdinet}.

\section{Preliminaries}
\subsection{Problem Formulation}
\label{Problem Formulation}

\begin{figure}[htbp]
\begin{center}
\centerline{\includegraphics[width=\columnwidth]{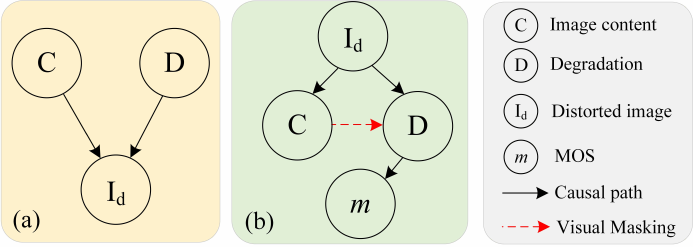}}
\caption{Structural Causal Model (SCM) for FR-IQA. Left: distorted image generation mechanism; Right: visual mechanism after decoupling.}
\label{SCM}
\end{center}
\end{figure}

As shown in Figure~\ref{SCM} (a), we construct a Structural Causal Model (SCM) for distorted image generation to describe the distorted image generation process. In this SCM, $C$ and $D$ respectively denote the image content and degradation. The path ${\rm{C}} \to {{\rm{I}}_d} \leftarrow {\rm{D}}$ indicates that the distorted image ${{\rm{I}}_d}$ is generated from both image content $C$ and degradation $D$. This SCM is based on the following assumption:
\begin{assumption}
\label{pro:factors}
A distorted image $I_d$ can be predominantly described by two factors: (1) image content $C$, which represents the semantic and structural elements, and (2) degradation $D$, which represents all distortions and quality degradation affecting the image.
\end{assumption}
\textbf{Assumption~\ref{pro:factors}} is a reasonable formulation within the \textbf{FR-IQA framework}. By definition, the reference image $I_r$ is distortion-free, thereby providing the pristine image content $C$. Accordingly, the distorted image $I_d$ is formulated as a combination of this image content, which strictly corresponds to $I_r$, and an externally applied degradation $D$. This process aligns with the causal generation path $C \to I_d \leftarrow D$.

\textbf{Building upon this generation mechanism and \textbf{Assumption~\ref{pro:factors}}, can we reformulate the FR-IQA task as a decoupling process?} FR-IQA assesses image quality by comparing a distorted image with its reference image, where the two images differ only in the introduced degradation. An intuitive idea is to decouple the degradation $D$ from $I_d$ using the pristine content $C$ provided by $I_r$, and predict the MOS based on the degradation $D$. However, this direct decoupling has a critical limitation, as relying only on the isolated degradation $D$ cannot accurately reflect human visual perception because image content can mask or accentuate distortions through the visual masking effect.

\begin{figure}[htbp]
\begin{center}
\centerline{\includegraphics[width=0.84\columnwidth]{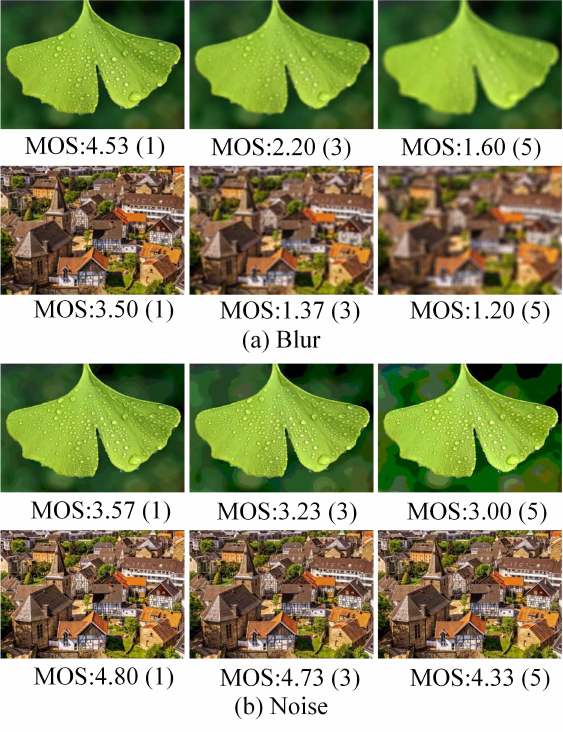}}
\caption{Illustrations of different distortion visibility. Under the same distortion
intensity level, the image with smooth content can be more tolerant to lens blur
than the image with textured content, yet more sensitive to quantization noise.
The value in parentheses indicates the imposed distortion intensity level.}
\label{vms}
\end{center}
\end{figure}

The principle of visual masking indicates that image quality is mainly related to the visibility of distortion~\cite{wang2023visual, zheng2024cdinet}. Specifically, the same image can exhibit different perceptual quality when it is affected by distortions of different types and intensity levels. However, the visibility of an identical distortion varies across different image contents. As shown in Figure~\ref{vms} (a), for lens blur, images with smooth content have higher MOS values than those with complex textures under the same distortion level, indicating that smooth content has a stronger masking ability for this distortion. However, images with smooth content are more sensitive to quantization noise than those with complex textures (see Figure~\ref{vms} (b)).

Therefore, although the degradation $D$ acts as the fundamental determinant of quality assessment, it is modulated by the image content $C$ due to the visual masking effect. Motivated by this insight, FR-IQA is reformulated as a process of causal disentanglement representation learning, as shown in Figure~\ref{SCM} (b). First, degradation features are decoupled from $I_d$ under the guidance of content invariance between the reference and distorted images. Second, the content-dependent modulation of degradation visibility, $C \to D$, is modeled. Finally, the resulting content-influenced degradation feature is used to predict the MOS.

\begin{figure*}[htbp]
	\centering
	\includegraphics[width=0.95\textwidth]{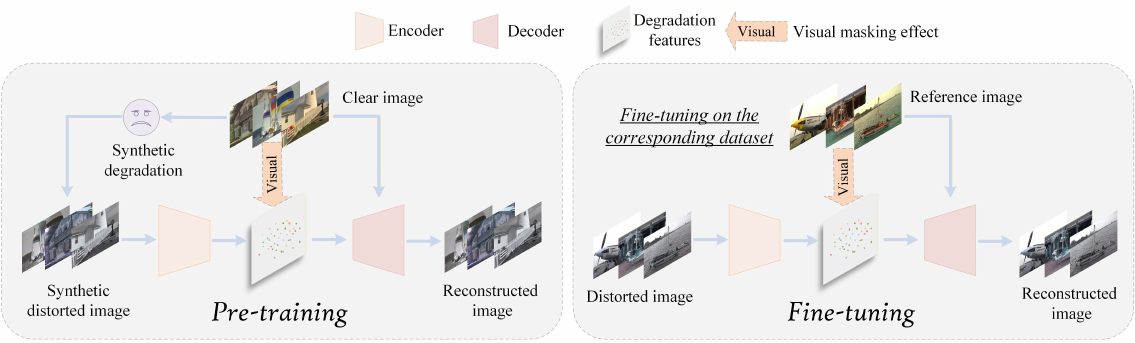}
	\caption{
    Overview of the pre-training and fine-tuning pipelines. Synthetic degraded pairs are constructed for reconstruction-based pre-training, and the same architecture is used for target-dataset fine-tuning.
    }
	\label{overall}
\end{figure*}

\subsection{Causal Disentanglement Representation Learning}
\label{Causal Disentanglement}

Causal disentangled representation learning extends conventional disentangled representation learning by considering structural dependencies among latent factors. Conventional methods usually assume that the underlying factors of variation are mutually independent and aim to separate them into distinct latent variables with clear semantic meanings~\cite{ICA, kingma2013auto, higgins2017beta}. However, this independence assumption can be too restrictive for real-world data, where different generative factors may be causally related.

In contrast, causal disentangled representation learning assumes that latent factors are not only semantically separable, but also structurally dependent. Most existing studies follow a reconstruction-based learning paradigm, in which the observed data are reconstructed from latent variables. Specifically, different factors are separated in the latent space, and their directed influences or structural relations are further modeled using causal graphs or structural causal models~\cite{yang2021causalvae, nam2023scadi, shen2022weakly}. Therefore, the learned representation is expected to preserve both the semantic meanings of individual factors and the dependencies among them.

Formally, let $\epsilon \in \mathbb{R}^{n}$ denote independent exogenous variables, and $z \in \mathbb{R}^{n}$ denote endogenous latent variables with causal semantics. Given an adjacency matrix $A$ that encodes the causal relations among latent variables, a linear structural relation can be written as
\begin{equation}
z = A^{\top}z + \epsilon = (I - A^{\top})^{-1}\epsilon,
\end{equation}
where $I$ is the identity matrix, and $(I - A^{\top})$ is assumed to be invertible. This formulation indicates that endogenous latent variables are generated from independent exogenous variables through structural dependencies encoded by $A$.

Inspired by this idea, our method also learns factorized representations under a reconstruction-related framework. Specifically, we separate content and degradation factors in FR-IQA and further model their task-specific structural relation. In this sense, our method is consistent with the principle of causal disentangled representation learning: it does not merely separate latent factors, but also exploits the dependency between them for quality assessment.

\section{Method}
\label{Method}
The proposed framework consists of three stages. Figure~\ref{overall} shows the process of pre-training and fine-tuning. In the pre-training stage, we first construct a synthetic degraded dataset from clear images. Subsequently, the model is pretrained using the constructed synthetic degraded dataset. In the fine-tuning stage, we fine-tune the model on specific IQA datasets (e.g., TID2013~\cite{ponomarenko2015image}, LIVE~\cite{sheikh2006statistical}) or other datasets containing distorted images and reference images. The model used for fine-tuning remains identical to that used for pre-training, with the only difference being the dataset. \textbf{In the label-free setting, the framework proceeds solely with pre-training, without subsequent fine-tuning.} Finally, in the prediction stage, we provide multiple approaches (i.e., supervised, few-shot, and label-free) to predict the quality score.

\subsection{Decoupling Degradation Features}
\label{sec:Decouple}

\begin{figure}[htbp]
	\centering
	\includegraphics[width=0.4\textwidth]{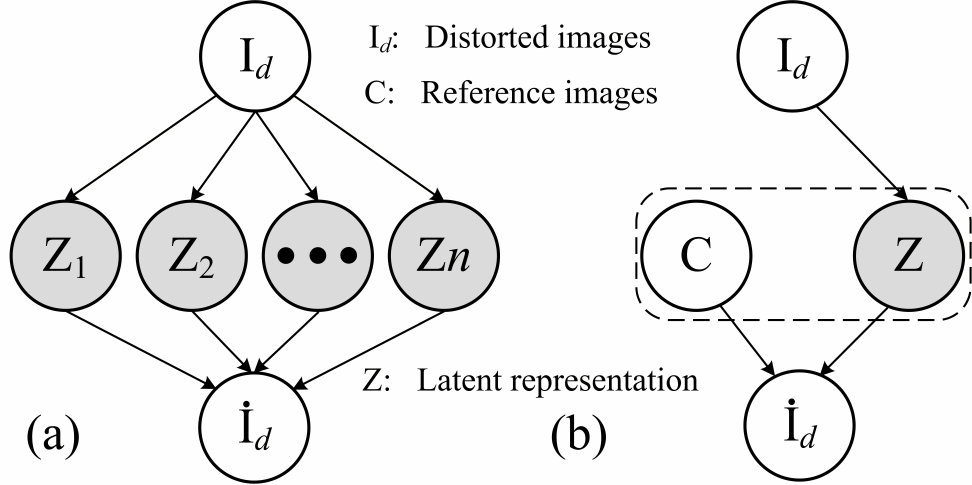}
	\caption{(a) The structure of the autoencoder. (b) With the reference image as a content condition, the latent representation is encouraged to encode degradation-related complementary information.}
	\label{fig:decouple}
\end{figure}

Guided by the problem formulation in Section~\ref{Problem Formulation}, our objective is to separate distortion-related factors from content-related factors in distorted images at this stage. We use an SCM to describe the latent generation process without intervention, as illustrated in Figure~\ref{fig:decouple} (a). The path \(I_d \rightarrow Z \rightarrow \dot I_d\) indicates that \(Z\) is the latent representation learned from the distorted image \(I_d\). Even if the autoencoder reconstructs the distorted image accurately, i.e., \(\dot I_d \approx I_d\), the latent representation may still remain entangled. Extracting decoupled degradation features with a standard autoencoder is underconstrained and requires additional inductive biases~\cite{locatello2019challenging}.

To support degradation separation, we exploit a task-specific property of FR-IQA. The reference image provides content information that is aligned with the distorted image. This property is stated as the following assumption.

\begin{assumption}[Content invariance in FR-IQA]
\label{ass:FR-IQA}
Under the standard FR-IQA setting considered in this work, the reference image \(I_r\) and the distorted image \(I_d\) share the same underlying scene content, while \(I_d\) additionally contains distortion-related factors.
\end{assumption}

Based on Assumption~\ref{pro:factors} and Assumption~\ref{ass:FR-IQA}, we use the reference image as an explicit content condition and learn a complementary latent representation from the distorted image. Specifically, the distorted image encoder extracts a latent representation
\begin{equation}
Z = E(I_d),
\label{eq:latent_feature}
\end{equation}
where \(E\) denotes the encoder.

The decoder reconstructs the distorted image by using the reference image as the content condition and \(Z\) as the complementary latent representation:
\begin{equation}
\dot I_d = G(I_r, Z)
          = G(I_r, E(I_d)),
\label{eq:content_conditioned_reconstruction}
\end{equation}
where \(G\) denotes the reconstruction decoder and \(\dot I_d\) is the reconstructed distorted image. The corresponding reconstruction objective is
\begin{equation}
\mathcal{L}_{\mathrm{rec}}
=
\mathbb{E}_{(I_r,I_d)}
\left[
d\!\left(G(I_r,E(I_d)), I_d\right)
\right],
\label{eq:content_conditioned_rec_loss}
\end{equation}
where \(d(\cdot,\cdot)\) denotes the reconstruction metric. 
This objective introduces a task-specific inductive bias for learning degradation-related representations. Since \(I_r\) already provides the aligned image content, the encoder \(E\) does not need to encode the complete content of \(I_d\). Instead, if the decoder can accurately reconstruct \(I_d\) from \(I_r\) and \(Z\), then the complementary information carried by \(Z\) should primarily correspond to the distortion-related changes that transform the reference image into the distorted image. Therefore, when Eq.~\eqref{eq:content_conditioned_rec_loss} is sufficiently minimized, Z tends to capture distortion-related residual information complementary to the reference content under the content invariance property of FR-IQA. In the following sections, we use \(D\) to denote the degradation-related latent representation.

\subsection{Content-Conditioned Visual Masking}
\label{Causal representation learning}

\begin{figure}[htbp]
    \centering
    \includegraphics[width=0.38\textwidth]{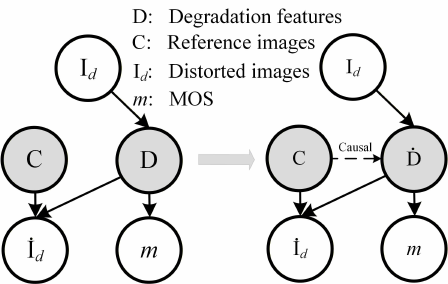}
    \caption{Causal disentanglement-inspired structural model for content-conditioned degradation perception.}
    \label{causalfig}
\end{figure}

Although the framework established in Section~\ref{sec:Decouple} can extract a degradation-related representation, directly using the decoupled degradation feature \(D\) remains insufficient for modeling human visual perception. In FR-IQA, the perceived severity of a distortion is determined not only by the degradation pattern, but also by the image content on which the degradation appears. This observation is consistent with the visual masking effect, in which image content may suppress or amplify the perceptual visibility of distortions.

We therefore introduce an effect learning formulation inspired by causal disentanglement to model content-conditioned degradation perception. As discussed in Section~\ref{Causal Disentanglement}, conventional disentangled representation learning usually encourages different latent factors to be statistically independent. In contrast, causal disentanglement representation learning further considers that separated latent factors may still have directed structural dependencies. Inspired by this view, we first treat the image content \(C\) and the degradation \(D\) as two separate factors, and then learn how the content factor modulates the perceptual expression of the degradation factor.

The proposed problem has a task-specific structure. In the FR-IQA setting, the reference image provides content information that is spatially aligned with the distorted image. Therefore, the latent space can be explicitly organized into two components, namely the image content \(C\) and the degradation representation \(D\). After the decoupling process, \(D\) mainly describes degradation-related information, while the content from the reference image provides a perceptual condition for degradation visibility.

Based on this task-specific structure, we introduce a directed structural dependency \(C \to {D}\) as a perceptually motivated structural prior, where image content regulates the perceptual visibility of degradation. Accordingly, the content-conditioned degradation representation is formulated as
\begin{equation}
\dot{D}=V(f_c(C),D),
\label{cau2}
\end{equation}
where \(f_c(C)\) denotes the content-related latent representation extracted from the reference image, and \(V(\cdot)\) denotes a learnable approximation of the visual masking effect, which transforms the decoupled degradation feature \(D\) into a content-influenced degradation feature \(\dot{D}\). This formulation enables the model to preserve the degradation representation while learning how its perceptual strength varies across different content conditions.

The proposed formulation incorporates the directed relation \(C \to {D}\) as a structural inductive bias inspired by causal disentanglement. This design follows the perceptual assumption that image content modulates the extent to which each degradation component is perceived under a specific content condition. As shown in Figure~\ref{causalfig}, the final distorted image reconstruction and quality prediction are therefore based on the image content \(C\) and the content-influenced degradation representation \(\dot{D}\). The detailed model design, including the encoder, visual masking module, and decoder, is presented in Section~\ref{Model_tructure}. The theoretical analysis is provided in the \textbf{Appendix II.A}.

\subsection{Score Prediction}
\label{predict}

\begin{figure*}[htbp]
	\centering
	\includegraphics[width=1\textwidth]{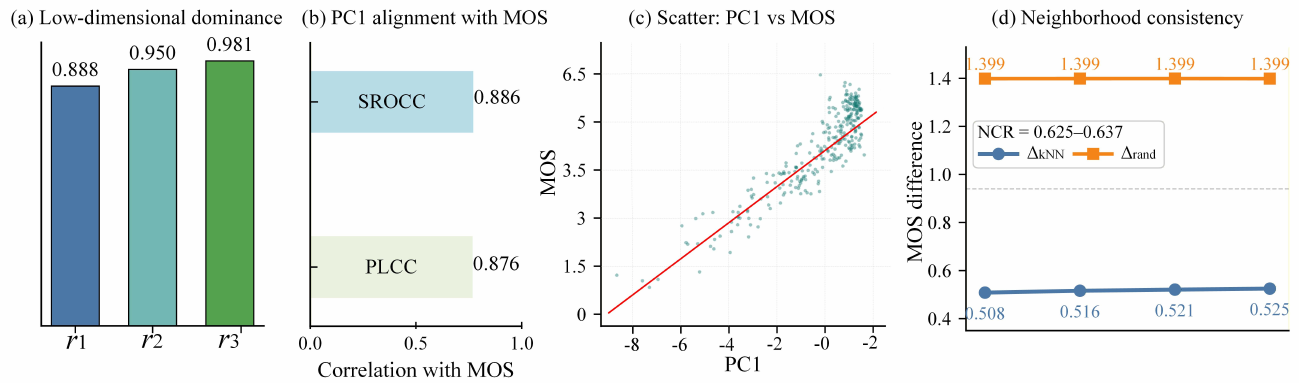}
	\caption{(a) Cumulative explained variance of the first three principal components. (b) Correlations between PC1 and MOS. (c) Scatter plot showing the relationship between PC1 and MOS. (d) Neighbors-quality consistency measured by the average MOS difference for k-nearest and random neighbors.}
	\label{iq_ana}
\end{figure*}

The final objective is to predict the image quality score based on the learned content-influenced degradation features $\dot{D}$. Depending on the availability of labeled IQA data, two complementary prediction strategies are introduced.

\subsubsection{Supervised regression} When labeled IQA data are available, we employ a two-layer fully connected (FC) regression head to map the content-influenced degradation features $\dot{D}$ to the final quality scores. During this phase, all parameters of the model are jointly updated in an end-to-end manner. This supervised training further aligns the learned content-influenced degradation representation with subjective human perception.

\subsubsection{Dimensionality reduction}
When labeled IQA data are unavailable, absolute MOS prediction is not identifiable. However, we observe that the perceptual representations used by existing FR-IQA methods contain a dominant principal component that is closely correlated with MOS. This observation suggests that the learned quality-related features can be transformed into a relative quality coordinate through dimensionality reduction without labeled data for training.

Specifically, for each distorted-reference image pair from LIVE, TID2013, and KADID, we perform quality prediction using four representative deep FR-IQA methods: WaDIQaM-FR~\cite{wadiqa}, IQT~\cite{iqt}, TOPIQ-FR~\cite{chen2024topiq}, and AHIQ~\cite{ahiq}. These methods are selected because they follow a paradigm similar to ours, in which the final quality score is predicted by mapping a feature representation to the score space. The quality score predicted by the $s$-th method can be written as  
\begin{equation}
\hat{m}_i^{(s)} = g^{(s)}\left(h_i^{(s)}\right),
\end{equation}
where \(g^{(s)}(\cdot)\) denotes the feature-to-score mapping of the \(s\)-th method and \(\hat{m}_i^{(s)}\) is the predicted quality score. \textbf{Since these FR-IQA methods are trained with MOS supervision, \(h_i^{(s)}\) is expected to encode quality-related information that is aligned with subjective perception.}

For each FR-IQA method, principal component analysis and kNN neighborhood analysis are performed on the normalized feature set \(\{h_i^{(s)}\}_{i=1}^{N}\). Before the analysis, each feature dimension is standardized to have zero mean and unit variance, which reduces the effect of feature-scale differences. We first examine whether the global variation of the feature space is dominated by low-dimensional components related to quality. Then, whether the local neighborhood structure in the feature space is consistent with the proximity structure induced by mean opinion scores (MOS) is evaluated. To reduce model-specific bias, the analysis is conducted independently for each selected FR-IQA method, and the average statistics over the four methods are reported.

As shown in Figure~\ref{iq_ana}, the first principal component explains \(88.8\%\) of the total variance, while the first two and three principal components explain \(95.0\%\) and \(98.1\%\), respectively. This result indicates that the variation of the extracted feature is highly concentrated along a dominant direction. More importantly, this direction is strongly correlated with MOS, achieving an SROCC of \(0.886\) and a PLCC of \(0.876\). Therefore, the dominant direction is not merely a high-variance direction, but is closely associated with subjective perceptual quality.

We further evaluate the consistency between feature-space neighborhoods and MOS proximity. For each sample, its nearest neighbors are identified in the feature space, and the average MOS difference between the sample and these neighbors is computed. As a random baseline, we select the same number of samples randomly and compute the corresponding MOS difference. When \(k\) is increased from \(5\) to \(20\), the kNN MOS difference is only increased from \(0.508\) to \(0.525\), whereas the random baseline remains around \(1.399\). This result indicates that samples close to each other in the learned feature space also tend to have similar subjective quality scores.

These results indicate that quality-aware representations contain a dominant low-dimensional structure that is highly correlated with subjective quality. This observation provides empirical motivation for estimating a relative quality coordinate from learned degradation-aware features.

Based on these observations, we adopt the following approximate model. Let \(h \in \mathbb{R}^{p}\) denote a quality-related feature. If the dominant quality variation of \(h\) is governed by a scalar coordinate \(q \in \mathcal{Q} \subset \mathbb{R}\), then \(h\) can be approximated as
\begin{equation}
h = \Gamma(q) + \eta,
\end{equation}
where \(\Gamma: \mathcal{Q} \rightarrow \mathbb{R}^{p}\) maps the scalar quality coordinate to the feature space, and \(\eta\) denotes residual variations that are not explained by this dominant coordinate. The subjective quality score is related to the same coordinate by
\begin{equation}
m = \psi(q) + \epsilon,
\end{equation}
where \(\psi(\cdot)\) is an unknown monotone mapping and \(\epsilon\) denotes subjective scoring noise with \(\mathbb{E}[\epsilon \mid q]=0\). Thus,
\begin{equation}
\mathbb{E}[m \mid q] = \psi(q).
\end{equation}

\begin{figure}[htbp]
    \centering
    \includegraphics[width=7.5cm]{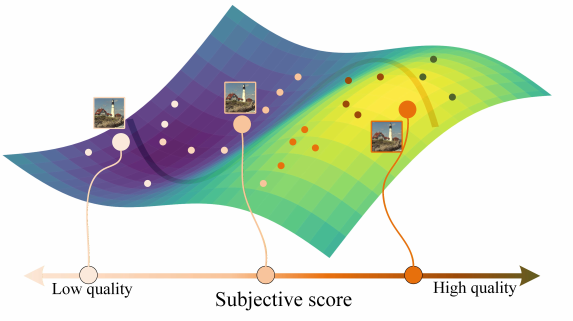}
    \caption{
    Relations in the feature space reflect the proximity of the latent perceptual quality values.}
    \label{manifold_projection}
\end{figure}

This formulation motivates our label-free score prediction strategy. As shown in Figure~\ref{manifold_projection}, in our framework, the prediction feature is the content-influenced degradation representation \(\dot{D}_i\). If \(\dot{D}_i\) sufficiently encodes perceptual quality, its MOS-related variation can be approximated by a scalar latent quality coordinate. Therefore, when MOS labels are unavailable, we estimate this coordinate directly from \(\dot{D}_i\) using unsupervised dimensionality reduction:
\begin{equation}
\tilde{q}_i = \Phi(\dot{D}_i),
\end{equation}
where \(\Phi(\cdot)\) denotes PCA, UMAP, or other dimensionality reduction functions. If \(\tilde{q}_i\) approximates a monotone transformation of \(q_i\), it preserves the relative quality coordinate. Therefore, in the label-free setting, the effectiveness of \(\tilde{q}_i\) is evaluated using correlation-based metrics, such as SROCC and PLCC.

\begin{figure}[htbp]
	\centering
	\includegraphics[width=8.5cm]{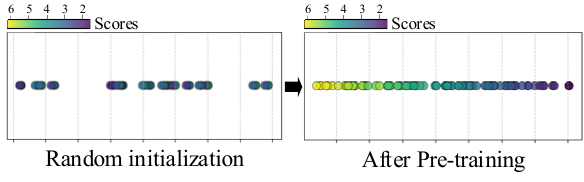}
	\caption{
      Visualization of the one-dimensional UMAP embedding on TID2013. The color of each point becomes darker as the ground-truth quality decreases. After pre-training, the samples are continuously distributed along the one-dimensional space according to quality variation, while random initialization fails to produce a discriminative structure.}
	\label{1dscat}
\end{figure}

As shown in Figure~\ref{1dscat}, the projected features are highly mixed in the one-dimensional space with random initialization, indicating that \(\dot{D}\) has not yet formed a quality-related structure. After pre-training, the projected samples exhibit a continuous distribution along the quality dimension, where the high-quality and low-quality samples become more separable. This result suggests that pre-training encourages the encoder and VMM to organize \(\dot{D}\) into a quality-aware representation, so that a relative quality coordinate can be recovered without MOS supervision.

\subsection{Model Structure}
\label{Model_tructure}
Figure~\ref{overall} shows the overall process, including degradation extraction, visual masking modulation, and degraded image reconstruction. The encoder first extracts the degradation feature \(D\) from the distorted image \(I_d\). The VMM then modulates \(D\) using the content representation extracted from the reference image \(I_r\). Finally, the decoder reconstructs \(I_d\) from \(I_r\) and the content-influenced degradation feature \(\dot{D}\).

\subsubsection{Encoder} The degradation encoder \(E\) is implemented using VGG16 and extracts the degradation representation from the distorted image. 

\subsubsection{Visual masking module} 
\begin{figure}[htbp]
    \centering
    \includegraphics[width=9cm]{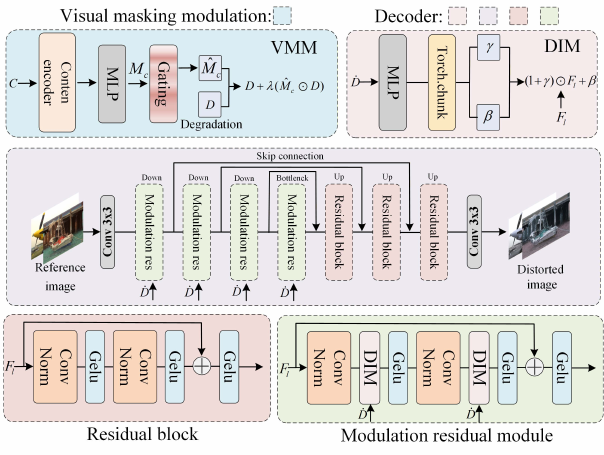}
    \caption{Architecture of the decoder and VMM.}
    \label{decoder}
\vspace{-1em}
\end{figure}

As shown in Figure~\ref{decoder}, the VMM captures the content-dependent masking effect in FR-IQA by explicitly accounting for how image content modulates distortion visibility. This design is necessary because the same physical distortion can exhibit different perceptual visibility across image contents. Therefore, the raw degradation feature D alone is insufficient for perceptual quality modeling, as it does not encode how image content affects the visibility of distortions.

Given the reference image \(I_r\), we use a separate content encoder \(f_c\), implemented using a ResNet18~\cite{he2016deep} backbone, to obtain the content representation
\begin{equation}
z_c = f_c(I_r).
\end{equation}
This content encoder is introduced to provide a clean content condition for VMM. Although the reconstruction network also extracts features from the reference image, these features are progressively fused with degradation information during the reconstruction process after \(\dot{D}\) is injected. Therefore, the content-conditioned mask is estimated from the independent reference representation \(z_c\), rather than from the reconstruction features. This design preserves the separation between content-conditioned modulation and degradation-conditioned reconstruction.

A nonlinear mapping \(f_m\), implemented with three fully connected layers, predicts a content-dependent modulation mask from \(z_c\):
\begin{equation}
M_c = \tanh(f_m(z_c)).
\end{equation}
To control the contribution of different degradation channels under the same content condition, another mapping \(f_g\) generates a channel gate:
\begin{equation}
a_c = \sigma(f_g(z_c)),
\end{equation}
where \(\sigma(\cdot)\) denotes the sigmoid function. The effective masking signal is defined as
\begin{equation}
\widehat{M}_c = a_c \odot M_c,
\end{equation}
where \(\odot\) denotes element-wise multiplication. Both \(M_c\) and \(a_c\) are projected to the same dimension as the degradation feature \(D\).

Given the degradation feature \(D\), the VMM produces a content-influenced degradation representation through residual modulation:
\begin{equation}
\dot{D} = D + \lambda \left(\widehat{M}_c \odot D\right),
\end{equation}
where \(\lambda\) controls the modulation strength. The residual form preserves the degradation representation itself and only changes its perceptual expression under a specific content condition. In this way, the VMM uses the reference content to selectively suppress or enhance degradation responses according to visual masking. The resulting representation \(\dot{D}\) therefore represents content-influenced degradation and is more consistent with human subjective perception than the raw degradation feature \(D\).

\subsubsection{Decoder} 
The decoder reconstructs the distorted image from the reference image and the content-influenced degradation feature. This reconstruction objective provides a self-supervised signal for learning degradation representations. Formally, the reconstruction process is written as
\begin{equation}
\dot{I}_d = \mathrm{Decoder}(I_r,\dot{D}),
\end{equation}
where \(I_r\) denotes the reference image, \(\dot{D}\) denotes the content-influenced degradation feature, and \(\dot{I}_d\) denotes the reconstructed distorted image.

The reconstruction network adopts a U-Net architecture. The reference image is first mapped into shallow features by a convolutional stem and then passed through an encoder to extract multi-scale reconstruction features. The output of the encoder at layer \(l\) is denoted as \(E_l\). These features provide structural information for reconstructing the distorted image and are passed to the corresponding decoder layers through skip connections to preserve spatial structures and fine details.

To inject degradation information into the reconstruction process, we design a degradation injection module (DIM). The DIMs in the decoder are conditioned only on \(\dot{D}\) and serve to guide image reconstruction. Thus, the independent content encoder provides the clean content condition for visual masking modulation, while the U-Net pathway provides reconstruction features from \(I_r\). The content-influenced degradation feature \(\dot{D}\) specifies the degradation pattern to be reconstructed.

For an intermediate feature \(F_l \in \mathbb{R}^{C_l \times H_l \times W_l}\), an MLP maps \(\dot{D}\) to the layer-specific channel-wise scaling and shifting parameters:
\begin{equation}
[\gamma_l,\beta_l] = \mathrm{MLP}_l(\dot{D}),
\end{equation}
where \(\gamma_l,\beta_l \in \mathbb{R}^{C_l}\). The DIM is applied as
\begin{equation}
\mathrm{DIM}(F_l,\dot{D}) = (1+\gamma_l)\odot F_l+\beta_l.
\label{eq:film}
\end{equation}
Here, \(\odot\) denotes channel-wise multiplication with broadcasting along the spatial dimensions. The term \(1+\gamma_l\) keeps the modulation close to an identity mapping at the early training stage and improves training stability. We apply the DIM in the downsampling layers and the bottleneck of the reconstruction U-Net.

Overall, the VMM performs perceptual modulation by converting the raw degradation feature \(D\) into the content-influenced degradation feature \(\dot{D}\) using an independent reference-only content representation. The decoder then uses \(\dot{D}\) as a degradation condition to reconstruct the distorted image from the reference image. Therefore, the VMM models content-conditioned distortion visibility, whereas the DIM implements degradation-conditioned reconstruction. The architecture of the decoder is illustrated in Figure~\ref{decoder}, and Figure~\ref{overall} shows the overall process. The \textbf{Appendix II.A} provides a theoretical analysis of how reconstruction encourages degradation-related learning and how the VMM performs bounded content-conditioned modulation.

\subsubsection{Loss function} The quality of the reconstructed image $\dot{\mathrm{I}}_{d}$ reflects the representation capacity of the extracted latent features. To comprehensively supervise this process, our objective function integrates three complementary components:

(i) \textbf{Pixel-wise Loss:} We employ Mean Squared Error (MSE) between the reconstructed output $\dot{\mathrm{I}}_{d}$ and the target distorted image $I_d$ to enforce basic pixel-level semantic and structural alignment:\begin{equation}\label{MSE}
{{\cal L}_{MSE}} = \frac{1}{{n \times m}}\sum\limits_{i = 1}^n {\sum\limits_{j = 1}^m {{{\left( {{{\mathop I\limits^. }_d}(i,j) - {I_d}(i,j)} \right)}^2}} } .
\end{equation}

(ii) \textbf{Perceptual Loss:} To mitigate the over-smoothing issue inherent to MSE~\cite{johnson2016perceptual}, we introduce a VGG19-based perceptual loss~\cite{ledig2017photo, simonyan2014very} to explicitly recover high-frequency structural details:
\begin{equation}
\label{vgg}
\mathcal{L}_{VGG} = \text{MSE}\left(\text{VGG}(\dot{I_d}), \text{VGG}(I_d)\right).\end{equation}

(iii) \textbf{Adversarial Loss:} Finally, a PatchGAN discriminator is utilized to synthesize realistic local textures and fine-grained visual details~\cite{patchgan}:
\begin{equation}\label{gan}\mathcal{L}_{GAN} = \log \left(\text{GAN}(I_d)\right) + \log \left(1 - \text{GAN}(\dot{I_d})\right).
\end{equation}

Finally, the above losses are combined to form the overall training objective:
\begin{equation}
\label{total_loss}
\mathcal{L}_{total}
=
\lambda_{MSE}\mathcal{L}_{MSE}
+
\lambda_{VGG}\mathcal{L}_{VGG}
+
\lambda_{GAN}\mathcal{L}_{GAN},
\end{equation}
where $\lambda_{\mathrm{MSE}}$, $\lambda_{\mathrm{VGG}}$, and $\lambda_{\mathrm{GAN}}$ denote the weighting coefficients for the pixel-wise loss, the VGG-based perceptual loss, and the adversarial loss, respectively. In our experiments, we set $\lambda_{\mathrm{MSE}}=1$, $\lambda_{\mathrm{VGG}}=1$, and $\lambda_{\mathrm{GAN}}=0.005$. This loss configuration follows common practice in GAN-based super-resolution and image restoration methods~\cite{20232degqe, patchgan}, where the adversarial term is assigned a relatively small weight to improve perceptual realism while maintaining reconstruction fidelity.

\subsection{Connection to the HVS and Existing Works}

\subsubsection{Connections to the HVS} The proposed causal disentanglement framework is closely related to the HVS. In human visual perception, the visibility of a distortion is strongly influenced by the spatial background, which corresponds to the psychophysical phenomenon known as the visual masking effect~\cite{ahumada1992luminance, pelli2013measuring, legge1980contrast, chen2010perceptually}. For instance, human eyes are significantly less sensitive to noise embedded in highly textured regions than in smooth and flat areas. Our framework emulates this human visual mechanism through causal disentanglement representation learning. Specifically, instead of extracting a monolithic feature in which image content and distortions are mixed, our SCM assumes that they can be represented as independent variables. Subsequently, by introducing the directed causal link $C \rightarrow D$, we provide a structured approximation of the masking effect. This allows the network to learn how semantic and structural information modulates the visibility of distortion, ensuring that the final representation $\dot{D}$ is better aligned with distortion visibility.

\subsubsection{Connections to existing models} Existing FR-IQA models typically employ dual-branch architectures to compute feature differences or similarities between the reference and distorted images. These methods are highly effective at quantifying signal fidelity by directly comparing the paired inputs. In this work, we propose a novel and complementary perspective by formulating FR-IQA as a causal disentanglement problem. Rather than focusing primarily on pairwise feature comparisons, the visual masking effect is explicitly modeled. This paradigm allows the modulation of distortion perception by image content to be encoded in a manner consistent with human perception.

\section{Experiments}
\label{Experiments}

\subsection{Experimental Protocols and Implementation Details}
\label{Experimental details}

\subsubsection{Training data and implementation details}
For standard IQA benchmarks, the synthetic pre-training set is constructed from high-quality pristine images in the Waterloo Database~\cite{Waterloo}. To match the degradation characteristics of each benchmark, we synthesize distorted images according to the degradation types and levels defined in the corresponding dataset protocol. For example, for LIVE, the five degradation types and levels specified in its protocol are generated. For domain-specific datasets, such as infrared or screen-content images, the synthetic pre-training data are instead constructed from pristine images in the corresponding domain, using the degradation types described in each dataset protocol. The pristine images used for domain-specific pre-training are strictly separated from the test images.

All input images are resized to \(448 \times 448\) pixels, randomly flipped with a probability of 0.5, and normalized. We use AdamW~\cite{loshchilov2017decoupled} as the optimizer, with an initial learning rate of \(1\times10^{-4}\) and a cosine annealing scheduler. The numbers of epochs for pre-training and fine-tuning are set to 200 and 20, respectively. The batch size is set to 24. The degradation encoder produces a 512-dimensional degradation embedding. The content encoder outputs a 256-dimensional content representation, which is fed into a two-layer MLP to generate a 512-dimensional content-conditioned mask. The modulation strength \(\lambda\) in the VMM is set to 0.3. In the reconstruction decoder, the DIM takes the 512-dimensional degradation embedding.  Fine-tuning uses the same reconstruction-based objective and network architecture as pre-training, with the only difference being the training dataset. 

\subsubsection{Training and prediction protocol}
The proposed framework is evaluated under three settings. When MOS labels are available, the model is first pretrained on the synthetic degraded dataset constructed from Waterloo and then fine-tuned on the target IQA dataset using paired reference and distorted images. This fine-tuning stage does not use MOS labels. Afterward, a two-layer MLP regression head is trained with MOS labels to map the learned content-modulated degradation feature \(\dot{D}\) to quality scores.

When MOS labels are unavailable on standard IQA benchmarks, the model is only pretrained on the Waterloo-based synthetic degraded dataset. The trained encoder and VMM are then frozen, and the content-influenced degradation representation \(\dot{D}\) is extracted from each evaluation sample. A dimensionality reduction method is further applied to project the high-dimensional features onto a one-dimensional quality coordinate, which is used as the relative quality score.

For domain-specific datasets, the model is pretrained on the corresponding domain-specific synthetic degraded dataset without using MOS labels. The trained encoder and VMM are then frozen, and quality scores are predicted from the extracted \(\dot{D}\) using the same dimensionality-reduction-based prediction strategy.

\subsubsection{Datasets}
We evaluate the proposed method on 10 IQA datasets, including five datasets that contain a wide range of distortion types (LIVE IQA~\cite{sheikh2006statistical}, CSIQ~\cite{larson2010most}, TID2013~\cite{ponomarenko2015image}, KADID~\cite{lin2019kadid}, and PIPAL~\cite{jinjin2020pipal}) and five domain-specific IQA datasets (infrared images~\cite{zelmati2022study}, radiographic images~\cite{zhang2025comprehensive}, screen-content images~\cite{ni2017scid}, medical images~\cite{medical}, and synthetic images~\cite{Synthetic}).

\begin{table*}[htbp]
\caption{Results of the performance comparison conducted on benchmark datasets.}
\label{zong}
\centering
\renewcommand{\arraystretch}{1.2}
\setlength{\tabcolsep}{1.8mm}

\begin{tabular}{c l *{10}{c}}
\toprule
\multirow{2}{*}{Category} & \multirow{2}{*}{Method}
& \multicolumn{2}{c}{LIVE~\cite{sheikh2006statistical}}
& \multicolumn{2}{c}{CSIQ~\cite{larson2010most}}
& \multicolumn{2}{c}{TID2013~\cite{ponomarenko2015image}}
& \multicolumn{2}{c}{KADID~\cite{lin2019kadid}}
& \multicolumn{2}{c}{PIPAL~\cite{jinjin2020pipal}} \\
\cmidrule(lr){3-4} \cmidrule(lr){5-6} \cmidrule(lr){7-8} \cmidrule(lr){9-10} \cmidrule(lr){11-12}
& & PLCC & SROCC & PLCC & SROCC & PLCC & SROCC & PLCC & SROCC & PLCC & SROCC \\
\midrule

& PSNR & 0.781 & 0.801 & 0.792 & 0.807 & 0.664 & 0.687 & 0.670 & 0.676 & 0.398 & 0.392 \\
\rowcolor{gray!20}\cellcolor{white}
& SSIM~\cite{wang2004image} & 0.847 & 0.851 & 0.810 & 0.833 & 0.665 & 0.627 & 0.610 & 0.619 & 0.489 & 0.486 \\
& MS-SSIM~\cite{wang2003multiscale} & 0.886 & 0.903 & 0.875 & 0.879 & 0.842 & 0.786 & 0.824 & 0.826 & 0.571 & 0.545 \\
\rowcolor{gray!20}\cellcolor{white}
& VIF~\cite{sheikh2006image} & 0.949 & 0.953 & 0.899 & 0.899 & 0.771 & 0.677 & 0.685 & 0.679 & 0.572 & 0.545 \\
& VSI~\cite{zhang2014vsi} & 0.877 & 0.899 & 0.912 & 0.929 & 0.898 & \textcolor{green}{0.895} & 0.877 & 0.878 & 0.548 & 0.526 \\
\rowcolor{gray!20}\cellcolor{white}
& GMSD~\cite{xue2013gradient} & 0.909 & 0.910 & 0.938 & 0.939 & 0.858 & 0.804 & 0.847 & 0.847 & 0.614 & 0.569 \\
\multirow{-7}{*}{Traditional}
& NLPD~\cite{laparra2016perceptual} & 0.882 & 0.889 & 0.913 & 0.926 & 0.832 & 0.799 & 0.809 & 0.812 & 0.489 & 0.464 \\

\midrule

\rowcolor{gray!20}\cellcolor{white}
& PieAPP (CVPR18)~\cite{prashnani2018pieapp} & 0.866 & 0.865 & 0.864 & 0.883 & 0.809 & 0.844 & 0.857 & 0.865 & 0.702 & 0.701 \\
& LPIPS (CVPR18)~\cite{zhang2018unreasonable} & 0.934 & 0.932 & 0.894 & 0.876 & 0.732 & 0.670 & 0.700 & 0.720 & 0.611 & 0.573 \\
\rowcolor{gray!20}\cellcolor{white}
& DISTS (TPAMI20)~\cite{ding2020image} & 0.924 & 0.925 & 0.919 & 0.920 & 0.854 & 0.830 & 0.886* & 0.886* & 0.645 & 0.627 \\
& PDL (ICCP21)~\cite{PDL} & 0.946 & 0.943 & 0.919 & 0.926 & 0.848 & 0.843 & 0.825 & 0.829 & 0.564 & 0.554 \\
\rowcolor{gray!20}\cellcolor{white}
& ADISTS (MM21)~\cite{ADISTS} & 0.954 & 0.955 & 0.957 & 0.950 & 0.858 & 0.834 & 0.889 & 0.889 & 0.645 & 0.631 \\
& DeepWSD (MM22)~\cite{liao2022deepwsd} & 0.904 & 0.925 & 0.941 & 0.950 & 0.894 & 0.874 & 0.887 & 0.888 & 0.503 & 0.500 \\
\rowcolor{gray!20}\cellcolor{white}
& DSD (NeurIPS21)~\cite{DSD} & 0.944 & 0.922 & 0.925 & 0.925 & 0.759 & 0.734 & 0.847 & 0.846 & 0.621 & 0.624 \\
& TOPIQ-FR (TIP24)~\cite{chen2024topiq} & 0.882 & 0.887 & 0.894 & 0.894 & 0.854 & 0.820 & 0.896 & 0.895 & \textcolor{red}{0.837*} & \textcolor{red}{0.809*} \\
\rowcolor{gray!20}\cellcolor{white}
& DeepJSD (TIP24)~\cite{DeepJSD} & \textcolor{red}{0.972} & \textcolor{green}{0.965} & \textcolor{green}{0.963} & \textcolor{red}{0.967} & 0.900 & 0.879 & 0.893 & 0.894 & 0.612 & 0.598 \\
& DeepCausal (CVPR25)~\cite{shen2025image} & 0.929 & 0.932 & 0.949 & 0.952 & \textcolor{blue}{0.909} & 0.884 & 0.898 & \textcolor{blue}{0.899} & 0.675 & 0.657 \\
\rowcolor{gray!20}\cellcolor{white}
& DeepDC (TIP25)~\cite{DeepDC} & 0.904 & 0.954 & 0.941 & \textcolor{green}{0.957} & 0.849 & 0.844 & 0.896 & \textcolor{red}{0.905} & 0.669 & 0.670 \\
\multirow{-12}{*}{Feature comparison}
& DBIQA (TPAMI25)~\cite{liao2025image} & \textcolor{green}{0.969} & \textcolor{blue}{0.962} & \textcolor{blue}{0.961} & \textcolor{blue}{0.956} & \textcolor{blue}{0.909} & \textcolor{green}{0.895} & \textcolor{green}{0.900} & \textcolor{green}{0.901} & \textcolor{green}{0.745*} & \textcolor{green}{0.769*} \\

\midrule

\rowcolor{gray!20}\cellcolor{white}
& Ours \textsuperscript{f}
& \textcolor{blue}{0.968} & \textcolor{red}{0.966} & \textcolor{red}{0.966} & 0.954 & \textcolor{red}{0.916} & \textcolor{red}{0.899} & \textcolor{red}{0.902} & 0.897 & \textcolor{blue}{0.727\textsuperscript{f}} & \textcolor{blue}{0.713\textsuperscript{f}} \\
& Ours w/ UMAP & 0.949 & 0.958 & 0.946 & 0.931 & \textcolor{blue}{0.909} & \textcolor{blue}{0.892} & \textcolor{blue}{0.899} & 0.891 & 0.684 & 0.666 \\
\rowcolor{gray!20}\cellcolor{white}
& Ours w/ PCA & 0.945 & 0.954 & 0.948 & 0.929 & \textcolor{green}{0.913} & \textcolor{green}{0.895} & 0.896 & 0.892 & 0.681 & 0.662 \\
& Ours w/ t-SNE & 0.919 & 0.923 & 0.921 & 0.928 & 0.906 & 0.881 & 0.874 & 0.883 & 0.665 & 0.653 \\
\rowcolor{gray!20}\cellcolor{white}
\multirow{-5}{*}{The proposed}
& Ours w/ TSVD & 0.942 & 0.948 & 0.933 & 0.922 & 0.901 & 0.882 & 0.893 & 0.885 & 0.673 & 0.652 \\

\bottomrule
\end{tabular}
\vspace{-1.5em}
\end{table*}

\subsection{Quality Prediction Performance}

In this study, we evaluate a variety of representative and state-of-the-art image quality assessment techniques on five IQA datasets, including LIVE~\cite{sheikh2006statistical}, CSIQ~\cite{larson2010most}, TID2013~\cite{ponomarenko2015image}, KADID~\cite{lin2019kadid}, and PIPAL~\cite{jinjin2020pipal}. The compared methods include PSNR, SSIM~\cite{wang2004image}, MS-SSIM~\cite{wang2003multiscale}, VIF~\cite{sheikh2006image}, VSI~\cite{zhang2014vsi}, GMSD~\cite{xue2013gradient}, NLPD~\cite{laparra2016perceptual}, PieAPP~\cite{prashnani2018pieapp}, LPIPS~\cite{zhang2018unreasonable}, DISTS~\cite{ding2020image}, PDL~\cite{PDL}, ADISTS~\cite{ADISTS}, DeepWSD~\cite{liao2022deepwsd}, DSD~\cite{DSD}, TOPIQ-FR~\cite{chen2024topiq}, DeepCausal~\cite{shen2025image}, DeepJSD~\cite{DeepJSD}, DeepDC~\cite{DeepDC}, and DBIQA~\cite{liao2025image}. These methods can be categorized into three groups: traditional IQA methods, feature comparison-based deep models, and the proposed causal disentanglement-based method. For the feature comparison-based deep models, we use the officially released model weights and evaluate their performance under the same testing protocol. For the traditional methods, we directly test them on each dataset.

Methods marked with an asterisk (*) were trained on external IQA datasets, including KADID and PIPAL. PieAPP and LPIPS were trained on their own proposed datasets. For the proposed method, we evaluate two prediction settings, including supervised fine-tuning and a fully unsupervised setting. Specifically, in the supervised fine-tuning setting \(^{f}\), the model is first pretrained on a synthetic dataset, then fine-tuned for feature extraction on a standard IQA dataset, and finally used to train an MLP-based regression model with labeled IQA data for prediction. For fair comparison, we fine-tune the model and perform supervised regression on the PIPAL dataset for cross-dataset testing. However, PIPAL~\cite{jinjin2020pipal} differs from conventional synthetic-distortion datasets because it contains not only traditional low-level distortions, but also outputs produced by image restoration and GAN-based algorithms. For many of these algorithm-induced distortions, the exact generation settings, model checkpoints, and degradation parameters are not fully specified in the public dataset release. Therefore, our synthetic pre-training protocol can only approximate part of the PIPAL distortion distribution, rather than exactly reproducing all its restoration- and GAN-related distortions. In the label-free setting, we only pretrain the model on the synthetic dataset and then use different dimensionality reduction methods to predict quality scores from the learned representations. Among these methods, UMAP and PCA generally achieve better and more stable performance than t-SNE and TSVD. This indicates that the learned content-influenced degradation representations contain a dominant quality-related structure that can be captured by both nonlinear manifold projection and linear principal directions. In contrast, t-SNE is more sensitive to local neighborhood preservation, while TSVD provides a weaker approximation of the perceptual quality axis. These results support the feasibility of deriving relative quality scores from the low-dimensional structure of the learned representation without MOS supervision.
\subsection{Domain-specific Dataset Experiments}

\begin{figure}[htbp]
	\centering
	\includegraphics[width=8cm]{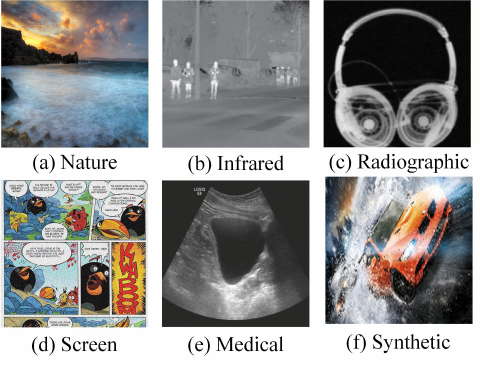}
	\caption{
      Image examples from diverse domains. From subfigure (a) to (f), there are natural image, infrared image, radiographic image, screen-content image, medical image, and synthetic image, respectively.}
	\label{domain}
\end{figure}

In some domain-specific image scenarios (e.g., infrared, radiographic, screen-content or medical radiographic imaging), the image structure and texture differ significantly from standard natural images. Figure~\ref{domain} shows images from different specific domains. Consequently, the existing state-of-the-art training-free methods that rely on models pretrained on the standard natural-image dataset ImageNet are often not suitable for these scenarios. Meanwhile, obtaining realistically degraded images and annotating their quality scores in these specialized domains is notoriously difficult. Therefore, the commonly used datasets typically consist of subjectively annotated synthetic distorted images~\cite{Synthetic,zelmati2022study}. However, constructing such IQA datasets requires a large number of expert evaluators, and the annotation process is extremely resource-intensive. 
The proposed method eliminates the dependence on MOS labels during adaptation.

\begin{table*}[htbp]
\caption{Results of the performance comparison conducted on domain-specific IQA datasets.}
\label{cross_completed}
\centering
\renewcommand{\arraystretch}{1.2}

\resizebox{\textwidth}{!}{%
\begin{tabular}{c l *{10}{c}}
\toprule
\multirow{2}{*}{Category} & \multirow{2}{*}{Method}
& \multicolumn{2}{c}{Infrared~\cite{zelmati2022study}}
& \multicolumn{2}{c}{Radiographic~\cite{zhang2025comprehensive}}
& \multicolumn{2}{c}{Screen~\cite{ni2017scid}}
& \multicolumn{2}{c}{Medical~\cite{medical}}
& \multicolumn{2}{c}{Synthetic~\cite{Synthetic}} \\
\cmidrule(lr){3-4} \cmidrule(lr){5-6} \cmidrule(lr){7-8} \cmidrule(lr){9-10} \cmidrule(lr){11-12}
& & PLCC & SROCC & PLCC & SROCC & PLCC & SROCC & PLCC & SROCC & PLCC & SROCC \\
\midrule

& PSNR & 0.814 & 0.815 & 0.861 & 0.873 & 0.778 & 0.763 & 0.615 & 0.609 & 0.603 & 0.590 \\
\rowcolor{gray!20}\cellcolor{white}
& SSIM~\cite{wang2004image} & 0.802 & 0.803 & 0.870 & 0.861 & 0.726 & 0.714 & 0.609 & 0.608 & 0.531 & 0.542 \\
& MS-SSIM~\cite{wang2003multiscale} & 0.832 & 0.827 & 0.859 & 0.854 & 0.757 & 0.741 & 0.656 & 0.636 & 0.712 & 0.699 \\
\rowcolor{gray!20}\cellcolor{white}
& VIF~\cite{sheikh2006image} & 0.907 & 0.901 & 0.833 & 0.842 & 0.819 & 0.796 & 0.591 & 0.583 & 0.748 & 0.755 \\
& VSI~\cite{zhang2014vsi} & 0.838 & 0.841 & 0.825 & 0.812 & 0.769 & 0.762 & 0.671 & 0.654 & 0.873 & 0.872 \\
\rowcolor{gray!20}\cellcolor{white}
\multirow{-6}{*}{Traditional}
& GMSD~\cite{xue2013gradient} & 0.889 & 0.891 & 0.938 & \textcolor{blue}{0.939} & 0.762 & 0.813 & 0.668 & 0.662 & 0.890 & 0.892 \\

\midrule

\rowcolor{gray!20}\cellcolor{white}
& LPIPS~\cite{zhang2018unreasonable} & 0.841 & 0.852 & 0.762 & 0.771 & 0.694 & 0.689 & 0.724 & 0.737 & 0.882 & 0.885 \\
& PDL~\cite{PDL} & 0.795 & 0.781 & 0.752 & 0.761 & 0.756 & 0.738 & 0.699 & 0.682 & 0.845 & 0.832 \\
\rowcolor{gray!20}\cellcolor{white}
& ADISTS~\cite{ADISTS} & 0.832 & 0.845 & 0.836 & 0.828 & 0.786 & 0.783 & 0.744 & 0.752 & 0.906 & 0.899 \\
& DeepWSD~\cite{liao2022deepwsd} & 0.862 & 0.858 & 0.876 & 0.885 & 0.880 & 0.799 & 0.808 & 0.784 & 0.902 & 0.883 \\
\rowcolor{gray!20}\cellcolor{white}
& DSD~\cite{DSD} & 0.813 & 0.824 & 0.839 & 0.827 & 0.830 & 0.829 & 0.762 & 0.771 & 0.876 & 0.870 \\
& DeepCausal~\cite{shen2025image} & 0.814 & 0.808 & 0.825 & 0.812 & 0.813 & 0.809 & 0.759 & 0.748 & 0.889 & 0.881 \\
\rowcolor{gray!20}\cellcolor{white}
& DeepJSD~\cite{DeepJSD} & 0.859 & 0.838 & 0.883 & 0.896 & 0.845 & 0.837 & 0.829 & 0.806 & 0.909 & 0.898 \\
\multirow{-8}{*}{Training-free}
& DeepDC~\cite{DeepDC} & 0.865 & 0.871 & 0.871 & 0.866 & 0.832 & 0.837 & 0.817 & 0.810 & 0.903 & 0.891 \\

\midrule

\rowcolor{gray!20}\cellcolor{white}
& Ours w/ UMAP
& 0.928 & \textcolor{blue}{0.921}
& \textcolor{blue}{0.943} & \textcolor{green}{0.940}
& \textcolor{blue}{0.905} & \textcolor{blue}{0.887}
& 0.852 & \textcolor{blue}{0.844}
& \textcolor{blue}{0.925} & \textcolor{blue}{0.921} \\
\multirow{-2}{*}{Generic Pre-training}
& Ours w/ PCA
& \textcolor{blue}{0.933} & 0.914
& 0.936 & 0.922
& 0.896 & 0.879
& \textcolor{blue}{0.858} & 0.835
& 0.918 & 0.914 \\

\midrule

\rowcolor{gray!20}\cellcolor{white}
& Ours w/ UMAP
& \textcolor{red}{0.956} & \textcolor{red}{0.948}
& \textcolor{green}{0.947} & \textcolor{red}{0.942}
& \textcolor{red}{0.926} & \textcolor{red}{0.915}
& \textcolor{green}{0.871} & \textcolor{red}{0.866}
& \textcolor{red}{0.947} & \textcolor{red}{0.951} \\
\multirow{-2}{*}{Specific Pre-training}
& Ours w/ PCA
& \textcolor{green}{0.952} & \textcolor{green}{0.941}
& \textcolor{red}{0.949} & 0.938
& \textcolor{green}{0.919} & \textcolor{green}{0.906}
& \textcolor{red}{0.875} & \textcolor{green}{0.861}
& \textcolor{green}{0.939} & \textcolor{green}{0.944} \\

\midrule
\end{tabular}%
}

\vspace{1mm}
\begin{minipage}{0.98\textwidth}
\footnotesize
\textit{Note:} Generic Pre-training denotes pre-training on Waterloo dataset. Specific Pre-training denotes pre-training with pristine images from the corresponding target domain.
\end{minipage}
\vspace{-1.5em}
\end{table*}

To validate the adaptability and effectiveness of our approach, we evaluate a variety of representative and state-of-the-art IQA techniques across five non-standard benchmark datasets: infrared images~\cite{zelmati2022study}, radiographic images~\cite{zhang2025comprehensive}, screen-content images~\cite{ni2017scid}, medical images~\cite{medical}, and synthetic images~\cite{Synthetic}. The compared methods include traditional metrics (PSNR, SSIM~\cite{wang2004image}, MS-SSIM~\cite{wang2003multiscale}, VIF~\cite{sheikh2006image}, VSI~\cite{zhang2014vsi}, GMSD~\cite{xue2013gradient}) and the existing state-of-the-art training-free deep feature approaches   (LPIPS~\cite{zhang2018unreasonable}, PDL~\cite{PDL}, ADISTS~\cite{ADISTS}, DeepWSD~\cite{liao2022deepwsd}, DSD~\cite{DSD}, DeepCausal~\cite{shen2025image}, DeepJSD~\cite{DeepJSD}, DeepDC~\cite{DeepDC}).

For the proposed method, we evaluate two label-free pre-training settings. In the generic pre-training setting, pristine images from the Waterloo dataset are used to synthesize degraded images according to the distortions of each target dataset. In the specific pre-training setting, pristine reference images from each target dataset are used instead, while the same target-dataset distortion protocol is applied. For fair evaluation, the pristine images used for specific pre-training are strictly separated from the test images. After pre-training, quality scores are predicted using UMAP or PCA.

As shown in Table~\ref{cross_completed}, the proposed method achieves strong performance on the evaluated domain-specific datasets under both generic and specific pre-training settings. The generic pre-training results show that the proposed method can generalize well even when the pristine images used for pre-training come from the Waterloo dataset rather than the target domains. This generalization ability can be attributed to the proposed strategy inspired by causal disentanglement. The proposed method first captures content-irrelevant degradation factors and then models their perceptual visibility conditioned on image content, rather than assessing distortions based on image statistical information. This design reduces the dependence on source-domain image distributions and thus improves cross-domain generalization.

Specific pre-training further improves the results on most datasets, indicating that target-domain pristine images provide additional domain-specific information for adaptation. Compared with existing training-free deep FR-IQA methods that rely on ImageNet-pretrained features, the proposed method shows stronger adaptability to domain-specific images without using any MOS labels.

\subsection{Controlled Validation of Degradation Disentanglement and Visual Masking}

To further examine whether the proposed method truly learns degradation-specific representations and whether the content-conditioned modulation partially captures distortion visibility consistent with visual masking, we design a controlled validation experiment to answer the following questions.

\noindent\textbf{Q1:} Can the proposed method disentangle degradation information from image content in the label-free setting?

\noindent\textbf{Q2:} Can the proposed method further model the visual masking effect in the label-free setting, such that the perceptual response to the same distortion varies adaptively with different image contents?

\begin{figure}[htbp]
	\centering
	\includegraphics[width=0.42\textwidth]{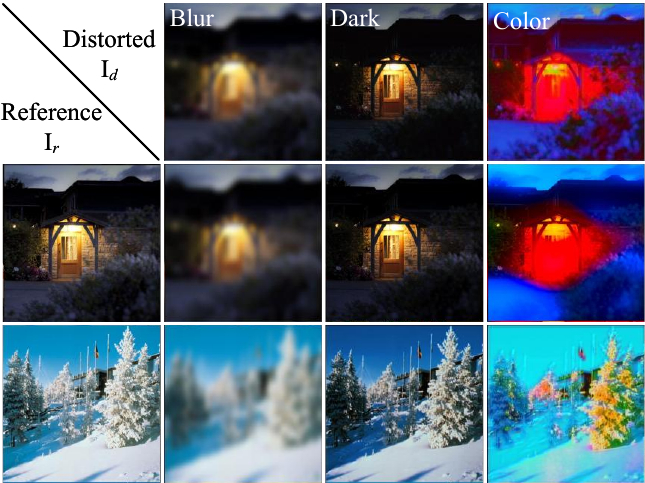}
	\caption{Counterfactual degradation transfer for answering Q1. The degradation feature $D_1$, extracted from a distorted image before the VMM, is combined with another reference image to examine whether the learned degradation representation is independent of image content.}
	\label{vis}
\end{figure}

\begin{figure}[htbp]
	\centering
	\includegraphics[width=0.5\textwidth]{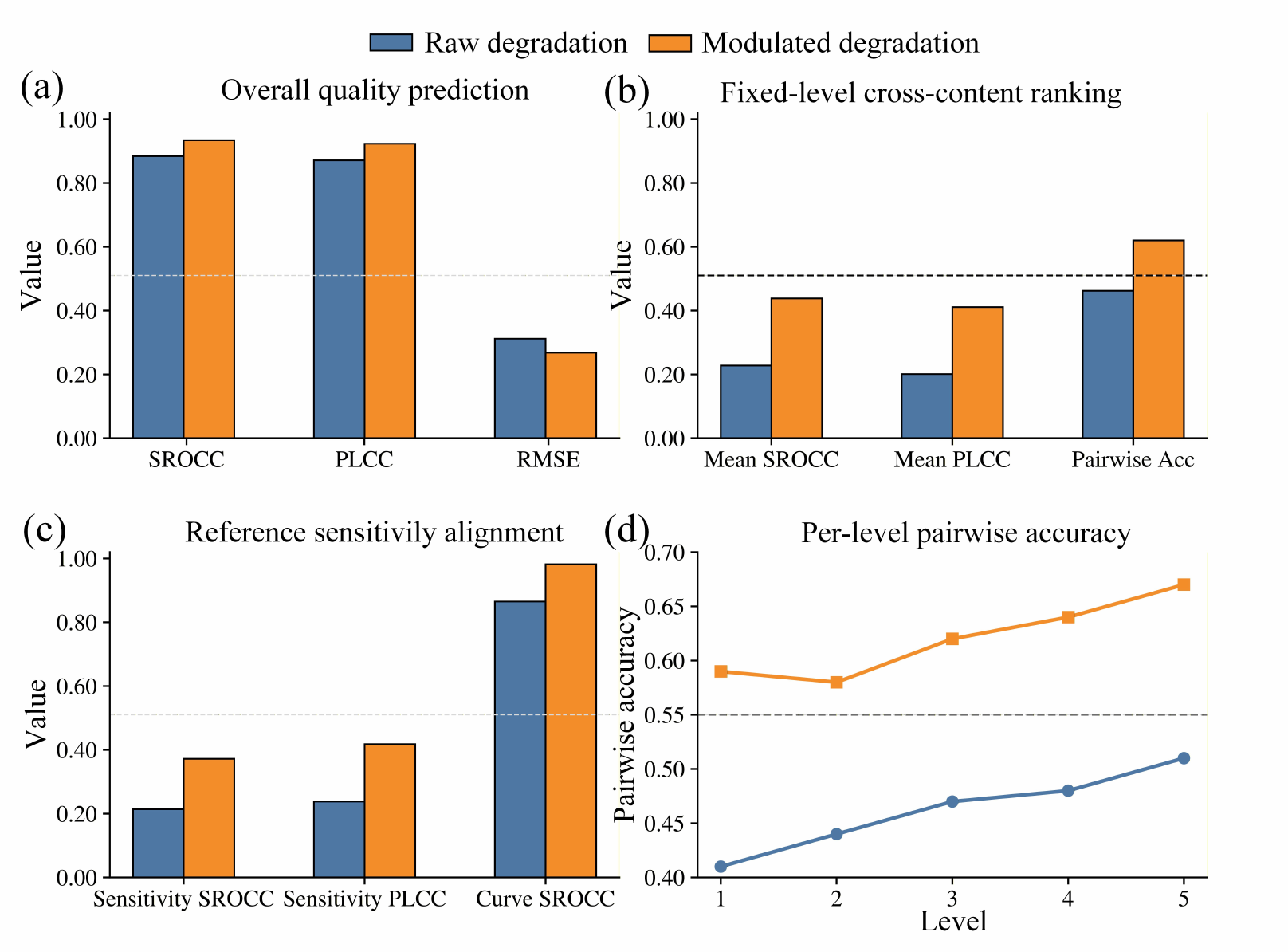}
	\caption{Controlled validation of visual masking for answering \textbf{Q2}. The raw degradation feature and the modulated degradation feature are compared on the TID2013 subsets. The evaluation includes overall quality prediction, fixed-level cross-content ranking, reference sensitivity alignment, and per-level pairwise accuracy.}
	\label{masking_validation}
\end{figure}

To answer \textbf{Q1}, we conduct a counterfactual degradation transfer experiment to verify whether the degradation representation extracted before the VMM is independent of image content. Given a distorted image, we extract its degradation representation $D_1$, where $D_1$ denotes the degradation feature before visual masking modulation. Then, $D_1$ is kept fixed, and another reference image $I_2$ is fed directly into the decoder together with $D_1$. The resulting image is generated as $\hat{I}_2^d = decoder({I_2},{D_1}).$

If \(D_1\) still contains strong content information from the source image, the decoder would introduce source-image content or unexpected structural changes when reconstructing \(\hat I_2^d\). In contrast, if \(D_1\) mainly describes degradation, the generated image should preserve the content of \(I_2\) while inheriting the distortion pattern from the source distorted image. As shown in Figure~\ref{vis}, the generated image largely keeps the content of the new reference image and transfers the corresponding degradation pattern. This result suggests that the representation before the VMM is mainly degradation-oriented and can be transferred across different image contents. \textbf{Appendix I.B} further validates the degradation-transfer property of raw \(D\) through quantitative PSNR/SSIM evaluation, and shows that directly transferring \(\dot{D}\) across different contents leads to reconstruction mismatch due to its content-conditioned nature.

To answer \textbf{Q2}, we conduct controlled experiments on the gaussian blur, noise, jp2k, color shift, contrast change, and dark subsets of TID2013 in the label-free setting. This experiment is designed to examine whether the VMM can modulate degradation responses according to image content under different distortion types. Specifically, we compare the raw degradation feature $D$, extracted before the VMM, with the modulated degradation feature $\dot D$, obtained after content-conditioned visual masking modulation. For each feature, a one-dimensional quality score is first derived and then compared with human subjective scores under controlled conditions to assess its consistency. All reported results are averaged over all subsets.

The two features are evaluated from four complementary aspects. First, overall quality prediction performance is measured on the distortion subsets to assess whether the learned feature is globally aligned with subjective quality. Second, fixed-level cross-content ranking is conducted to evaluate content-aware quality discrimination under identical distortion strengths. Specifically, within each distortion level, all distorted images share the same physical distortion strength but are generated from different reference contents. Ranking correlation and pairwise ranking accuracy are therefore computed within each distortion level, followed by averaging across all levels and distortion subsets. This metric evaluates whether the feature can distinguish content-dependent distortion visibility when physical distortion strength is controlled. Third, reference-level sensitivity alignment is evaluated. For each reference image, we fit the subjective quality scores across distortion levels as a function of the distortion level, and use the resulting slope to represent human sensitivity to the corresponding distortion. The same procedure is applied to the model responses, and the correlation between model sensitivity and human sensitivity is computed across reference images. Finally, per-level pairwise accuracy is reported to quantify how consistently the model preserves content-dependent perceptual ordering at each distortion level.

\begin{table*}[htbp] 
\centering
\caption{Few-shot MOS calibration results under different labeled data ratios.}
\label{tab:zero_four_datasets}
\renewcommand{\arraystretch}{1.2} 
\setlength{\tabcolsep}{3pt} 
\resizebox{\textwidth}{!}{ 
\begin{tabular}{c cccc cccc cccc cccc} 
\toprule
\multirow{3}{*}{Scale} & \multicolumn{4}{c}{\textbf{LIVE}} & \multicolumn{4}{c}{\textbf{TID2013}} & \multicolumn{4}{c}{\textbf{KADID}} & \multicolumn{4}{c}{\textbf{Radiographic}} \\
\cmidrule(lr){2-5} \cmidrule(lr){6-9} \cmidrule(lr){10-13} \cmidrule(lr){14-17}
& \multicolumn{2}{c}{Frozen} & \multicolumn{2}{c}{Unfrozen} & \multicolumn{2}{c}{Frozen} & \multicolumn{2}{c}{Unfrozen} & \multicolumn{2}{c}{Frozen} & \multicolumn{2}{c}{Unfrozen} & \multicolumn{2}{c}{Frozen} & \multicolumn{2}{c}{Unfrozen} \\
\cmidrule(lr){2-3} \cmidrule(lr){4-5} \cmidrule(lr){6-7} \cmidrule(lr){8-9} \cmidrule(lr){10-11} \cmidrule(lr){12-13} \cmidrule(lr){14-15} \cmidrule(lr){16-17}
& PLCC & SROCC & PLCC & SROCC & PLCC & SROCC & PLCC & SROCC & PLCC & SROCC & PLCC & SROCC & PLCC & SROCC & PLCC & SROCC \\
\midrule
0\%  & 0.949 & 0.958 & 0.949 & 0.958 & 0.909 & 0.892 & 0.909 & 0.892 & 0.899& 0.891 & 0.899& 0.891 & 0.947 & 0.942 & 0.947 & 0.942 \\
\rowcolor{gray!15}
2\%  & 0.932 & 0.921 & 0.805 & 0.809 & 0.895 & 0.876 & 0.768 & 0.752 & 0.884 & 0.872 & 0.745 & 0.738 & 0.928 & 0.918 & 0.801 & 0.796 \\
5\%  & 0.938 & 0.936 & 0.803 & 0.786 & 0.901 & 0.884 & 0.795 & 0.787 & 0.889 & 0.881 & 0.771 & 0.782 & 0.934 & 0.927 & 0.806 & 0.812 \\
\rowcolor{gray!15}
10\% & 0.940 & 0.939 & 0.895 & 0.881 & 0.907 & 0.890 & 0.853 & 0.839 & 0.894 & 0.888 & 0.842 & 0.835 & 0.939 & 0.935 & 0.889 & 0.878 \\
15\% & 0.943 & 0.941 & 0.912 & 0.908 & 0.911 & 0.895 & 0.878 & 0.864 & 0.898 & 0.893 & 0.869 & 0.861 & 0.942 & 0.939 & 0.908 & 0.901 \\
\bottomrule
\end{tabular}
}
\vspace{-1em}
\end{table*}

\begin{table*}[htbp]
\centering
\renewcommand{\arraystretch}{1.2} 

\begin{minipage}[t]{0.51\textwidth}
    \centering
    \caption{Ablation study on reconstruction losses.}
    \label{tab:loss}
    \resizebox{\linewidth}{!}{%
    \begin{tabular}{c c c  c c  c c  c c}
    \toprule
    \multicolumn{3}{c}{Loss} & \multicolumn{2}{c}{LIVE} & \multicolumn{2}{c}{TID2013} & \multicolumn{2}{c}{Radiographic}\\
    \cmidrule(lr){1-3} \cmidrule(lr){4-5} \cmidrule(lr){6-7} \cmidrule(lr){8-9}
    $\mathcal{L}_{MSE}$ & $\mathcal{L}_{VGG}$ & $\mathcal{L}_{GAN}$ & PLCC & SROCC & PLCC & SROCC & PLCC & SROCC  \\
    \midrule
    \checkmark & \xmark & \xmark     & 0.764 & 0.751 & 0.741 & 0.732 & 0.775 & 0.746 \\
    \rowcolor{gray!20}
    \checkmark & \checkmark & \xmark & 0.919 & 0.906 & 0.901 & 0.883 & 0.938 & 0.927 \\
    \checkmark & \xmark & \checkmark & 0.806 & 0.797 & 0.775 & 0.764 & 0.821 & 0.833 \\
    \rowcolor{gray!20}
    \xmark & \checkmark & \checkmark & 0.692 & 0.674 & 0.653 & 0.661 & 0.512 & 0.490 \\
    \checkmark & \checkmark & \checkmark & \textbf{0.949} & \textbf{0.958} & \textbf{0.909} & \textbf{0.892} & \textbf{0.947} & \textbf{0.942} \\
    \bottomrule
    \end{tabular}
    }
\end{minipage}
\hfill
\begin{minipage}[t]{0.45\textwidth}
    \centering
    \caption{Ablation study on training strategies.}
    \label{strategies}
    \resizebox{\linewidth}{!}{%
    \begin{tabular}{c c c  c c  c c  c c}
    \toprule
    \multicolumn{3}{c}{Method} & \multicolumn{2}{c}{LIVE} & \multicolumn{2}{c}{TID2013} & \multicolumn{2}{c}{Radiographic} \\
    \cmidrule(lr){1-3} \cmidrule(lr){4-5} \cmidrule(lr){6-7} \cmidrule(lr){8-9} 
    PT\vphantom{$\mathcal{L}_{MSE}$} & FT & VMM & PLCC & SROCC & PLCC & SROCC & PLCC & SROCC \\
    \midrule
    \checkmark & \xmark & \xmark     & 0.912 & 0.920 & 0.872 & 0.871 & 0.921 & 0.915\\
    \rowcolor{gray!20}
    \checkmark & \checkmark & \xmark & 0.931 & 0.929 & 0.901 & 0.891 & 0.944 & 0.946 \\
    \checkmark & \xmark & \checkmark & 0.949 & 0.958 & 0.909 & 0.892 & 0.947 & 0.942 \\
    \rowcolor{gray!20}
    \xmark & \checkmark & \checkmark & 0.183 & 0.162 & 0.201 & 0.198 & 0.748 & 0.698\\
    \checkmark & \checkmark & \checkmark & \textbf{0.968} & \textbf{0.966} & \textbf{0.916} & \textbf{0.899} & \textbf{0.958} & \textbf{0.949} \\
    \bottomrule
    \end{tabular}
    }
\end{minipage}
\vspace{-1em}
\end{table*}
As shown in Figure~\ref{masking_validation}, the modulated degradation feature consistently outperforms the raw degradation feature. On average over the TID2013 subsets, $\dot D$ improves SROCC from $0.884$ to $0.934$ and PLCC from $0.871$ to $0.923$, while reducing RMSE from $0.312$ to $0.268$. In the fixed-level cross-content ranking experiment, $\dot D$ improves the mean-level SROCC from $0.228$ to $0.438$, the mean-level PLCC from $0.201$ to $0.411$, and the pairwise accuracy from $0.462$ to $0.620$. Moreover, in reference sensitivity alignment, $\dot D$ improves the sensitivity SROCC from $0.214$ to $0.372$ and the sensitivity PLCC from $0.238$ to $0.418$, while increasing the mean curve SROCC from $0.928$ to $0.982$. Since the fixed-level ranking experiment controls the physical distortion strength, the improvement of $\dot D$ suggests that the proposed method can, to some extent, model the content-dependent visibility of distortions across different distortion types. These results suggest that the proposed modulation improves the alignment with content-dependent perceptual responses, which is consistent with the visual masking effect.

\subsection{Few-shot Experiments}

Unsupervised dimensionality reduction methods, such as UMAP, provide a relative quality coordinate that is expected to preserve the perceptual ordering of image quality, rather than an absolute MOS prediction calibrated to the score scale of each dataset. To further examine whether the learned representation can be calibrated to absolute subjective scores with limited supervision, we conduct few-shot MOS regression experiments on LIVE, TID2013, KADID, and Radiographic, as reported in Table~\ref{tab:zero_four_datasets}. The experiments are conducted with $0\%$, $2\%$, $5\%$, $10\%$, and $15\%$ labeled data. The $0\%$ setting corresponds to the label-free prediction protocol, while the remaining settings use a small subset of labeled samples to train a lightweight regression head. The rest of the data are used for evaluation.

We compare two training strategies. In the frozen setting, the degradation feature extractor is fixed and only the regression head is trained. The regression head consists of a normalization layer followed by a linear layer, which maps the learned content-influenced degradation representation to the MOS scale. In the unfrozen setting, both the regression head and the degradation feature extractor are updated using the available labeled samples. As shown in Table~\ref{tab:zero_four_datasets}, the frozen setting maintains consistently high PLCC and SROCC across different label ratios, indicating that the learned degradation representation is already strongly aligned with subjective quality and can be effectively calibrated with only a small number of labels. In contrast, updating the degradation feature extractor with very limited labeled data leads to a clear performance drop. This suggests that the small labeled subset is insufficient to reliably optimize the feature extractor and may instead disturb the representation learned from reconstruction-based pre-training. These results support the robustness of the proposed representation in both label-free and few-shot quality prediction scenarios.

\subsection{Further Analysis}

\subsubsection{The impact of loss} During pre-training and fine-tuning, we employ three loss functions, including MSE loss, VGG loss, and GAN loss. Pixel-wise losses such as MSE mainly constrain low-level intensity consistency, but they are less effective in modeling the uncertainty of lost high-frequency details, such as textures and fine structures. As shown in Table~\ref{tab:loss}, when MSE loss is only used, the model tends to produce over-smoothed reconstructions and shows limited sensitivity to high-frequency degradations. As a result, the learned representation is insufficient for accurately capturing degradation patterns dominated by high-frequency noise.
\begin{figure}[htbp]
	\centering
	\includegraphics[width=8cm]{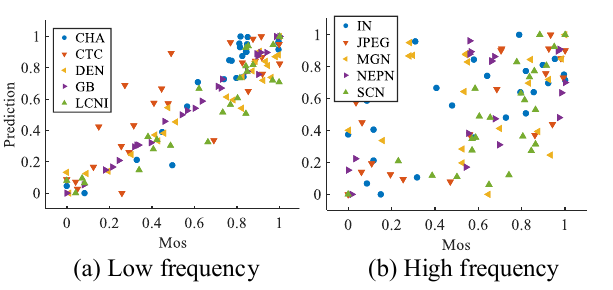}
	\caption{
       Prediction consistency under low-frequency and high-frequency distortions. Each point denotes one distorted image, where the horizontal axis is MOS and the vertical axis is the predicted score.}
	\label{hd}
\end{figure}
Introducing $\mathcal{L}_{VGG}$ on top of $\mathcal{L}_{MSE}$ consistently brings the most substantial improvement across the evaluated datasets. This result indicates that perceptual supervision plays a central role in learning degradation-aware representations. Compared with pure pixel-wise reconstruction, VGG loss encourages the model to preserve perceptually meaningful structures and texture-related information, which improves its ability to represent both low-frequency and high-frequency degradations. Figure~\ref{hd} further illustrates this effect. When the model is trained only with MSE loss, it can evaluate certain low-frequency degradation types reasonably well, but it fails to provide reliable assessment for high-frequency degradation types.

The adversarial loss further improves the fidelity of fine details and textures in the reconstructed images, thereby enhancing the representation capability of the model. The full loss combination achieves the best overall performance on all datasets. Compared with the improvement brought by $\mathcal{L}_{VGG}$, the gain from $\mathcal{L}_{GAN}$ is relatively smaller, which suggests that adversarial supervision mainly serves as a refinement term rather than the primary source of degradation-aware representation learning.

\subsubsection{The impact of training strategies} We investigate the contributions of three training strategies. Specifically, PT denotes synthetic degradation-based pre-training, FT denotes the combined stage of reconstruction-based target-dataset fine-tuning and MOS-supervised regression, and VMM denotes the proposed visual masking module. Even the use of only the pre-training scheme yields strong performance on the LIVE, TID2013, and Radiographic datasets. VMM captures content-dependent modulation of degradation features. With this layer, the learned representation is better aligned with human subjective perception in an unsupervised setting, leading to consistent performance gains across all three datasets. After labeled data are introduced during the final stage of FT, the model is more precisely aligned with human subjective perception and achieves the best performance on all three datasets.

\begin{table*}[htbp]
\centering
\renewcommand{\arraystretch}{1.2}
\setlength{\tabcolsep}{3pt}

\begin{minipage}[t]{0.50\textwidth}
    \vspace{0pt}
    \centering

    \begin{minipage}[t][3.0\baselineskip][t]{\linewidth}
        \centering
        \caption{Ablation study on different masking strategies in the VMM.}
        \label{tab:causal_masking_ablation}
    \end{minipage}
    \vspace{0.15cm}

    \resizebox{\linewidth}{!}{%
    \begin{tabular}{l c c c c c c c}
    \toprule
    \multirow{2}{*}{Masking Strategy} & \multirow{2}{*}{Formulation} 
    & \multicolumn{2}{c}{LIVE} 
    & \multicolumn{2}{c}{TID2013} 
    & \multicolumn{2}{c}{Radiographic} \\
    \cmidrule(lr){3-4} \cmidrule(lr){5-6} \cmidrule(lr){7-8}
    & & PLCC & SROCC & PLCC & SROCC & PLCC & SROCC \\
    \midrule
    None & $\dot{D}=D$ 
    & 0.912 & 0.920 & 0.872 & 0.871 & 0.921 & 0.915 \\
    \rowcolor{gray!20}
    Concatenation & $\dot{D}=\mathrm{MLP}([D;z_c])$ 
    & 0.923 & 0.938 & 0.895 & 0.878 & 0.925 & 0.921 \\
    Additive & $\dot{D}=D+\lambda m$ 
    & 0.922 & 0.931 & 0.894 & 0.883 & 0.932 & 0.928 \\
    \rowcolor{gray!20}
    Multiplicative & $\dot{D}=m\odot D$ 
    & 0.931 & 0.935 & 0.899 & 0.881 & 0.938 & 0.934 \\
    Residual Masking & $\dot{D}=D+\lambda(m\odot D)$ 
    & 0.944 & 0.953 & 0.906 & 0.890 & 0.943 & 0.939 \\
    \rowcolor{gray!20}
    Gated Residual Masking & $\dot{D}=D+\lambda(\hat{m}\odot D)$ 
    & \textbf{0.949} & \textbf{0.958} 
    & \textbf{0.909} & \textbf{0.892} 
    & \textbf{0.947} & \textbf{0.942} \\
    \bottomrule
    \end{tabular}%
    }
\end{minipage}
\hfill
\begin{minipage}[t]{0.48\textwidth}
    \vspace{0pt}
    \centering

    \begin{minipage}[t][3.0\baselineskip][t]{\linewidth}
        \centering
        \caption{Ablation study on the structures and the content encoder in the VMM.}
        \label{causal}
    \end{minipage}
    \vspace{0.15cm}

    \resizebox{\linewidth}{!}{%
    \begin{tabular}{l c c c c c c c}
    \toprule
    \multirow{2}{*}{Configuration} & \multirow{2}{*}{Causal Graph} 
    & \multicolumn{2}{c}{LIVE} 
    & \multicolumn{2}{c}{TID2013} 
    & \multicolumn{2}{c}{Radiographic} \\
    \cmidrule(lr){3-4} \cmidrule(lr){5-6} \cmidrule(lr){7-8}
    & & PLCC & SROCC & PLCC & SROCC & PLCC & SROCC \\
    \midrule
    Ours w/o VMM & $C\perp D$ 
    & 0.912 & 0.920 & 0.872 & 0.871 & 0.921 & 0.915 \\
    \rowcolor{gray!20}
    Inverse & $D\to C$ 
    & 0.908 & 0.916 & 0.865 & 0.869 & 0.917 & 0.914 \\
    CausalVAE & $C\to D$ 
    & 0.923 & 0.938 & 0.895 & 0.878 & 0.925 & 0.921 \\
    \rowcolor{gray!20}
    VGG16 & $C\to D$ 
    & 0.941 & 0.939 & 0.896 & 0.885 & 0.933 & 0.929 \\
    ResNet18 & $C\to D$ 
    & \textbf{0.949} & \textbf{0.958} 
    & \textbf{0.909} & \textbf{0.892} 
    & \textbf{0.947} & \textbf{0.942} \\
    \bottomrule
    \end{tabular}%
    }
\end{minipage}

\end{table*}

\subsubsection{The impact of VMM}

We evaluate several content-conditioned modulation strategies in the VMM. Specifically, direct concatenation is formulated as $\dot{D}=\mathrm{MLP}([D;z_c])$, additive modulation as $\dot{D}=D+\lambda m$, multiplicative masking as $\dot{D}=m\odot D$, and residual masking as $\dot{D}=D+\lambda(m\odot D)$. For the full gated residual masking used in our method, we further introduce a channel gate $\hat{m}=\sigma(g)\odot m$, and define the final modulation as $\dot{D}=D+\lambda(\hat{m}\odot D)$. This experiment is designed to verify whether image content should be directly fused with degradation features or should instead act as a variable that modulates the perceptual visibility of degradation components.

The results in Table~\ref{tab:causal_masking_ablation} validate the necessity of the VMM. Removing visual masking modulation leads to a substantial performance drop on all datasets, which indicates that modeling the content-dependent visibility of distortions is critical for learning perceptually meaningful degradation features. Direct concatenation improves upon the baseline but remains less effective than modulation-based formulations. This result suggests that directly combining content and degradation information is less effective than explicitly treating content as a modulation factor. Additive and multiplicative variants further improve performance, showing that content-aware adjustment is beneficial. However, pure multiplicative masking may over-suppress the original degradation signal. In contrast, residual masking preserves the original degradation representation while allowing content to regulate its perceptual expression, resulting in better performance. Finally, introducing an additional channel gate further improves the results, which demonstrates that the visibility modulation should itself be adaptively weighted across channels.

Table~\ref{causal} reports the ablation study on the VMM direction and the content encoder. Specifically, the causal structure between content and degradation is not interchangeable. Compared with the variant without causal modeling ($C \perp D$), our full model with $C \to D$ achieves clear and consistent gains on all datasets. In contrast, reversing the direction to $D \to C$ results in inferior performance, indicating that the proposed $C \to D$ formulation better matches the underlying dependency in distorted images. The choice of content encoder substantially affects performance. When the same causal graph ($C \to D$) is used, ResNet18 consistently surpasses VGG16 across LIVE, TID2013, and Radiographic, yielding the strongest PLCC and SROCC in every case. This demonstrates that the proposed VMM benefits from a more expressive feature extractor. The ResNet18/VGG16-based architecture also outperforms CausalVAE~\cite{yang2021causalvae}. This improvement likely stems from the difference in modeling focus: CausalVAE is designed for generic disentanglement of multiple latent factors, whereas our method explicitly models the content--degradation relationship that is central to IQA.

\subsubsection{Supplementary experiments} Additional experimental results are provided in the supplementary material due to space limitations. These include ablation studies on the degradation encoder, decoder modulation strategy, and masking formulation, together with degradation-transfer validation and additional visualizations (details in \textbf{Appendix I}). These experiments verify the robustness of the proposed method and show that the learned representation captures transferable degradation information while enabling perceptual modulation.

\section{Conclusions}
\label{Conclusions}

This paper presents a causal-disentanglement-inspired framework for FR-IQA. Different from existing deep FR-IQA methods that mainly estimate perceptual quality through pairwise feature comparison between reference and distorted images, the proposed framework reformulates degradation estimation as a process of content-degradation separation and content-conditioned degradation perception. By exploiting the content invariance between reference and distorted images, the model decouples degradation-related representations from image content without relying on MOS supervision. The proposed VMM further models the influence of image content on degradation visibility, which enables the extraction of content-influenced degradation representations that are more consistent with the visual masking effect. The learned representation supports both supervised and label-free quality prediction. When MOS annotations are available, a regression head can be used to map the learned representation to subjective quality scores. When annotations are unavailable, dimensionality reduction provides a one-dimensional quality coordinate for relative quality ranking. Extensive experiments on standard IQA benchmarks demonstrate that the proposed method achieves competitive performance under fully supervised, few-shot, and label-free settings. Experiments on infrared, radiographic, screen-content, medical, and synthetic image datasets further show that the method exhibits stronger label-free adaptation than existing training-free FR-IQA models. These results indicate that the proposed causal-disentanglement-inspired framework provides a promising alternative to conventional feature-comparison-based FR-IQA, reducing the dependence on subjective annotations while improving adaptability to non-standard and domain-specific image scenarios.

Future work can further improve the proposed framework in two directions. First, the current VMM is still an approximate model of content-conditioned distortion visibility and may not fully capture the complex perceptual characteristics of the HVS. Visual masking modules that better conform to human visual perception can be designed to improve the alignment between learned degradation representations and subjective perception. Second, more effective modules and supervision signals can be introduced to better reconstruct degraded images, while spatial alignment or geometry-aware reconstruction mechanisms can be incorporated to address geometric displacement between reference and distorted images. This may improve the robustness of the method against GAN-based and locally inconsistent distortions.


\bibliography{main}
\bibliographystyle{IEEEtran}

\newpage
\vfill
\end{document}